%% file: main.tex
\def\year{2021}\relax
\documentclass[letterpaper]{article} %
\usepackage{aaai21}  %
\usepackage{times}  %
\usepackage{helvet} %
\usepackage{courier}  %
\usepackage[hyphens]{url}  %
\usepackage{graphicx} %
\usepackage{natbib}  %
\usepackage{caption} %
\usepackage[utf8]{inputenc} %
\DeclareUnicodeCharacter{2212}{-}

\usepackage{microtype}
\usepackage{graphicx}
\usepackage{booktabs} %
\usepackage{multirow}
\usepackage{siunitx}
\usepackage[frozencache,cachedir=.]{minted}

\usepackage{listings}
\usepackage[gen]{eurosym} %
\usepackage{fancyhdr}
\usepackage{lastpage}

\usepackage[hyphens]{url}
\usepackage[]{hyperref}
\hypersetup{breaklinks=true}

\usepackage{pgfplots}
\pgfplotsset{compat=newest}
\usepgfplotslibrary{groupplots}
\usepgfplotslibrary{dateplot}

\usepackage{amsmath}
\usepackage{tikz}
\usepackage{mathdots}
\usepackage{yhmath}
\usepackage{cancel}
\usepackage{color}
\usepackage{siunitx}
\usepackage{array}
\usepackage{multirow}
\usepackage{amssymb}
\usepackage{gensymb}
\usepackage{tabularx}
\usepackage{booktabs}
\usetikzlibrary{fadings}
\usetikzlibrary{patterns}
\usetikzlibrary{shadows.blur}

\usepackage{wrapfig}

\usepackage{enumitem}
\setitemize{noitemsep,topsep=0pt,parsep=0pt,partopsep=0pt}

\newcommand{\bigO}{\mathcal{O}}

\usepackage{caption}
\usepackage{subcaption}

\usepackage{appendix}

\usepackage{amsmath,bm}
\usepackage{amssymb}

\usepackage[autostyle, english=american]{csquotes}
\MakeOuterQuote{"}

\everypar{\looseness=-1} %
\usepackage{adjustbox}

\usepackage{hyperref}

\newcommand\textcite{\citet}

\begin{document}

\title{Classifying Sequences of Extreme Length with Constant Memory\\Applied to Malware Detection}

\author {
        Edward Raff, \textsuperscript{\rm 1,2, 3} 
        William Fleshman, \textsuperscript{\rm 4} 
        Richard Zak, \textsuperscript{\rm 1,2, 3} \\
        Hyrum S. Anderson, \textsuperscript{\rm 5} 
        Bobby Filar, \textsuperscript{\rm 6} 
        Mark McLean \textsuperscript{\rm 1} 
        \\
}
\affiliations {
    \textsuperscript{\rm 1} Laboratory for Physical Sciences, 
    \textsuperscript{\rm 2} Booz Allen Hamilton, 
    \textsuperscript{\rm 3} University of Maryland, Baltimore County \\
    \textsuperscript{\rm 4} U.S. Army, 
    \textsuperscript{\rm 5} Microsoft, 
    \textsuperscript{\rm 6} Elastic \\
}

\maketitle

\begin{abstract}
Recent works within machine learning have been tackling inputs of ever-increasing size, with cybersecurity presenting sequence classification problems of particularly extreme lengths. In the case of Windows executable malware detection, inputs may exceed $100$ MB, which corresponds to a time series with $T=100,000,000$ steps. To date, the closest approach to handling such a task is MalConv, a convolutional neural network capable of processing up to $T=2,000,000$ steps. The $\bigO(T)$ memory of CNNs has prevented further application of CNNs to malware. In this work, we develop a new approach to temporal max pooling that makes the required memory invariant to the sequence length $T$. This makes MalConv $116\times$ more memory efficient, and 
up to $25.8\times$
faster to train on its original dataset, while removing the input length restrictions to MalConv. We re-invest these gains into improving the MalConv architecture by developing a new Global Channel Gating design, giving us an attention mechanism capable of learning feature interactions across 100 million time steps in an efficient manner, a capability lacked by the original MalConv CNN. Our implementation can be found at
\url{https://github.com/NeuromorphicComputationResearchProgram/MalConv2}
\end{abstract}

\section{Introduction}
\label{sec:intro}

Cybersecurity has received increased attention from machine learning practitioners and researchers due to the number of challenges that exist within the space. Industry datasets are routinely measured in petabytes \cite{tagkey2014iv}, have noisy labels, are both structured and unstructured, suffer from continuous concept drift \cite{Kantchelian:2013:AAD:2517312.2517320}, and adversarial attacks have been well motivated as a daily occurrence for decades \cite{Rajab2011}. In this work we are interested in the task of \textit{static malware detection}, where using the on-disk byte representation, one wishes to predict if a new executable program is benign or malicious. Current industry models rely heavily on domain knowledge feature extraction, which is time consuming and expensive, and requires intimate knowledge of Windows and low-level assembly and software design. Because malware authors adapt, this feature engineering is a continuous processes, which can require reverse engineering effort to determine what new features should be extracted. 
To quantify the cost of such efforts, a single program can take weeks for experts  with decades of experience to reverse engineer \cite{Votipka2019}, so the ability to build models that perform their own feature extraction and construction can save an enormous amount of time if successful.  

Toward this goal, we follow the approach of MalConv, which proposed to tackle the problem of malware detection as a time series classification problem \cite{MalConv}. For an input file $x$ of $T$ bytes in length, a neural network must learn to produce an output label $y \in \{\text{Benign}, \text{Malicious}\}$. The MalConv architecture was relatively small ($\approx$ 1 million parameters) but represented a meaningful malware detector by performing convolutions over raw byte inputs. Additionally, the work identified a number of challenges with this task. In particular, their approach would process up to 2 MB of a file---equivalent to a time series prediction problem with $T=2,000,000$ steps. 
The next longest time series task we are aware of is only on the order of $\leq 16,000$ steps \cite{wavenet}.
Due to the extreme length of raw byte inputs, the MalConv solution required an NVIDIA DGX-1 with 128 GB of GPU memory to train over one month of compute time.  This has made MalConv difficult to replicate, while simultaneously neglecting the fact that 2 MB is relatively small with respect to the distribution of observed executable file sizes, where the tails can reach in excess of 100 MB. 

In this work, we produce a solution to the high memory cost to train MalConv, making the memory use \textit{invariant} to the length of the input --- allowing us to train on data points in excess of 200,000,000 time steps in length using a single GPU. 
This reduces the memory used to train MalConv by a factor of 116$\times$ while simultaneously providing an up to 25.8$\times$ speedup, reducing the compute requirements from a DGX-1 down to a free Google Colab instance. 
Our approach leverages the sparse gradients of temporal max pooling to cap the memory requirement during training on long inputs. By significantly reducing the runtime and memory constraints of MalConv, we are able to explore more advanced architectures for the task of time series classification. In particular, we develop a new \textit{global channel gating} (GCG) that allows us to enhance MalConv to learn interactions of features across the entire input space. GCG can be implemented in only 7 lines of Python, making it easy to implement while improving the accuracy of the end-to-end deep malware detection model. Although we explicitly address malware detection where long input sequences are dramatic, our contributions are relevant generally to deep neural networks with long input sequences, some of which are discussed in the following section. We also note the task of learning interactions over a sequence of unprecedented length of intrinsically interesting from a pure ML perspective and benefits a real-world task.  

We have organized the paper as follows. In \autoref{sec:related_work} we will review the work related to MalConv, which has received significant attention from the cybersecurity community that motivates our research, as well as other work in the domain of processing long input sequences. Next we will detail our approach to making the memory cost of MalConv style architectures invariant to feature length in \autoref{sec:fixed_memory}. These improvements are necessary to make our global channel gating possible, which we detail in \autoref{sec:global_channel_gating}. The results detailing our speedups and memory improvements are presented in \autoref{sec:results}, followed by our conclusions in \autoref{sec:conclusion}.

\section{Related Work} \label{sec:related_work}

The desire to perform malware classification from raw bytes, to alleviate expensive and constant feature engineering, has long existed. This was originally based on the Normalized Compression Distance (NCD) \cite{Li2004}, which has found extensive use for this task \cite{Wehner:2007:AWN:1370628.1370630,Bailey:2007:ACA:1776434.1776449,Hayes2008,Bayer2009,Borbely2015,Alshahwan2015,10.1145/3292006.3302385,Menendez2019,S.Resende2019,Walenstein2007}. Recent works like LZJD~\cite{raff_lzjd_2017,raff_lzjd_digest,pylzjd-proc-scipy-2019} and BWMD~\cite{Raff2020} are built from compression algorithms and useful in unsupervised settings, but are less effective in supervised ones. We will use these methods as baselines to compare against. 

MalConv was the first proposed approach to detect malware from raw bytes, processing inputs of up to 2 MB in length \cite{MalConv}. Through a broad search across network architectures, the authors report that many classical "best practices" for neural network architectures did not apply. 
For example, they found that BatchNorm prevented convergence, and that a network with 1 layer of extremely wide convolutions performed better than deeply stacked narrow filters.
Since \cite{MalConv},
a number of others have replicated their approach or proposed alterations to better study it, but all have reduced the input size in order to reduce computational costs. Authors from the anti-virus company Avast 
restricted their study to files that where $\leq 512$ KB \cite{Krcal2018}. Their work is notable for being the first to compare the approach with hand engineered domain knowledge features from their production malware classifier. They found that the CNN was close in performance, and combining the domain and CNN features improved accuracy by 4\%, indicating the CNN was learning features or feature interactions not previously found by domain experts. 
FireEye did an in-depth reverse engineering of what a MalConv-like network learned showing it corresponded well to what an analyst would look for, but had to restrict their model to 100 KB\cite{Coull2019}. 
\textcite{Anderson2018} introduced the Ember dataset and found MalConv slightly worse than hand-crafted features, but needed 25 hours/epoch to train on up-to 1 MB. 
Recent work\cite{Galinkin2019} has even shown MalConv has an ability to generalize across x86 architectures, detecting x86 macOS and Linux malware when trained only on Windows data.
Other works have used the same or similar architectures to perform malware detection on datasets other than Windows executables, including for Android APKs\cite{Hasegawa2018}, PDF files\cite{Jeong2019}, as well as malicious JavaScript and Visual Basic code detection\cite{Stokes2018}. 

These works have all demonstrated the value of the byte based approach to malware detection, but simultaneously show the computational limitations. These solutions all suffer from an artificial limit in the maximum file size imposed by memory constraints; these potentially degrade performance and enable easy evasion in an adversarial scenario.
Many works have shown MalConv is susceptible to evasion \cite{Demetrio2019,Kolosnjaji2018,Kreuk2018,Fleshman2018}, but these attacks can be thwarted at a cost to accuracy \cite{Fleshman2018a}.
This defense is only moderately effective because MalConv can be thwarted by simply inserting the malicious payload after the 2 MB file limit. 
Because malware authors are real active adversaries attempting to evade detection, this is a serious limitation. After years of activity and development, our work finally removes this trivial limitation from this research area, which also makes  \cite{Fleshman2018a} more effective. 
While MalConv has received significant interest for its applications in malware detection, few other works within machine learning approach the same length of sequence processing. Recent work extending the Transformer approach to more efficiently handle long inputs has reached $T=64,000$ time steps \cite{Kitaev2020}. While the Transformer is able to learn more robust representations than our current work, it is still orders of magnitude too short to be able to process most executable files. Work by \textcite{NIPS2019_9689} proposed an extension of Recurrent Neural Networks, showing them to be capable of learning on synthetic time series of $T=1,000,000$ steps. Their approach requires over an hour to process a single time series of this length, making it computationally infeasible --- where our approach enables MalConv to run on a similar length input in under 42 milliseconds. While our approach improves the representational power of MalConv and is faster to train, it has less representational power compared to these other works. We provide more details on the failed attempts with transformers, and other approaches in a "What Did Not Work" \autoref{sec:did_not_work}. 

Our approach to fixing the memory cost of MalConv is similar to checkpoint (or "rematerialization") \cite{10.1145/347837.347846}. This approach involves re-computing results during the backward pass to avoid saving results in the forward pass, trading more compute for less memory but guaranteeing identical results. All work in this domain has focused on ways to balance this trade off for different types of acyclic network graphs \cite{Chen2016,NIPS2016_6221,NIPS2019_9653,NIPS2019_8400,Beaumont2020}. Our work instead performs recomputation in the forward pass, so that the backward pass produces an equivalent result, while using less compute time and less memory. 

Although we focus exclusively on the application of malware detection from byte sequences, we note that other domains may similarly benefit from tools for classification over long time series. For example, Genome Wide Association Studies (GWAS) can exceed 500,000 time steps in length, and have long dealt with issues in discovering interactions across GWAS\cite{Wu2010}. When constrained to smaller sequences with $T \leq 5000$, architectures similar to MalConv have found use for GWAS based prediction tasks \cite{Liu2019}.

\section{Fixed Memory Convolution Over Time} \label{sec:fixed_memory}

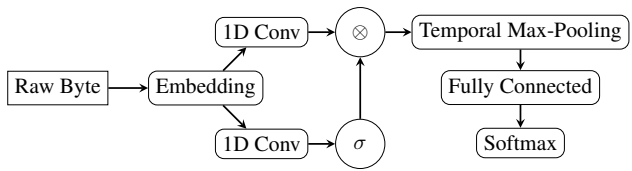
\begin{figure}[!htb]
    \centering
    \adjustbox{max width=\columnwidth}{%
        \input{figs/malconv.tex}
    }
    \caption{Original MalConv architecture~\cite{MalConv} with $\approx$ 1M parameters, but required 128 GB of GPU RAM to train. $\otimes$ indicates element-wise product, and $\sigma$ the sigmoid activation. }
    \label{fig:malconv}
\end{figure}

The original MalConv architecture is shown in \autoref{fig:malconv}. It contains an embedding layer (of $\mathbb{R}^8$) that is used over an alphabet of 257 tokens, 256 bytes + an End of File marker. These are fed into two sets of 128 convolutional filters with a width of 512 and a stride of 512\footnote{Originally a width and stride of 500 was used, but it has been noted in several works that using a power of two performs better due to assembly code being aligned on powers of two when written to an executable. }, which are then used in a gating approach proposed by \cite{pmlr-v70-dauphin17a}. The gated result is then converted to a fixed length feature vector using temporal max pooling (i.e., global max pooling, or max pooling over time), after which it is fed into a simple fully connected layer for prediction of the benign/malicious label. Since there is only one layer, the receptive window size $W$ is equal to the kernel width 512.

Despite its simplicity, MalConv was the first architecture to demonstrate that neural networks could learn to perform malware detection from raw bytes and the first to show classification over time series/sequences of up to $T=2,000,000$ steps. However, only the first 2 MB of the input was processed in training MalConv because it required 128 GB of GPU memory to train on a batch of 256 files up to the 2MB limit. This is owing to the large memory cost of performing an embedding and convolution over a time series of 2 million steps (1 for each byte), and the resulting activations alone require almost all of the GPU memory. Every subsequent work we are aware of has processed less than the original 2 MB cap.

To overcome these issues, we  developed a novel Temporal Max-Pooling strategy that makes memory costs invariant to the sequence length $T$.  Importantly, we do this by noting that Temporal Max-Pooling causes the gradient with respect to the sequence to be sparse. For $C$ channels, saving all $C \cdot T$ activations is unnecessary, as only $C$ values will actually be used, one for each channel. Thus we are using many times more memory than needed to train, and also performing redundant GPU computations on the backward pass since the majority of gradient values are exactly 0.  When working with normal images and standard applications of max-pooling, the sparsity ratio may be 1:2 or 1:4, which is generally not sparse enough to make exploitation of that fact useful. This is because every non-zero value requires storing its associated index, doubling the memory use of those values. Second, operations on dense vectors/matrices result in more efficient computation, delivering computational throughput closer to the theoretical limits of modern hardware when using modern BLAS libraries and software like CUDNN. As such, libraries like PyTorch
and Tensorflow
do not support sparse gradients through max-pooling.

\begin{figure}[!htb]
    \centering
    \adjustbox{width=\columnwidth}{%
        \input{figs/lowMemPool.tex}
    }
    \caption{Diagram of Temporal Max Pooling with fixed memory. The original input (top) is a 1D sequence with 3 channels and is broken up into four chunks based on window size $W=3$. Without gradient computation/tracking, the maximum activation index is found within each chunk. 
    Solid colors show max values kept, "$\times$" max in chunk but no maximal.
    Winning indices are copied to a new shorter sequence (bottom), which runs with gradient tracking.
    The result is the same output and gradient, but fixed memory cost. 
    }
    \label{fig:fixed_mem_conv}
\end{figure}
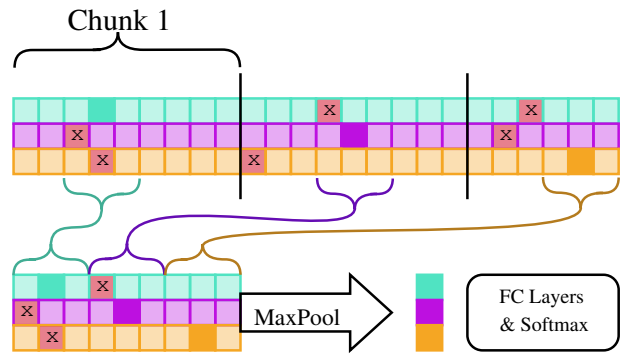

Conversely, we obtain the benefits of sparse activations while also retaining the higher computational throughput of dense operations, without requiring any new code, as follows. 
\begin{itemize}[topsep=0pt,parsep=0pt,partopsep=0pt,leftmargin=12pt]
    \item[1.] Turn off gradient computation (e.g., \mintinline{python}{with torch.no_grad():} if using PyTorch) and break the input sequence of length $T$ into at most $T/(W\cdot 2)$ \textit{overlapping} chunks of size $W\cdot 3$.
    \item[2.] Perform max pooling over each chunk; for each channel track the maximum value and its absolute index.
    \item[3.] Compare values within each chunk to a set of global winners. If a new chunk's maximal activation exceeds the global winner, it becomes the new global winner.
\end{itemize}
Once we have processed all chunks, we know the $C$ locations, one for each channel, that will win the max-pooling over time. The chunks overlap so that this computation is correct, and not impacted by windowing issues. 

With these $C$ locations, we may simply concatenate their values into a new sequence of length $T'=C\cdot W$. This new sequence is now small enough that the full set of embedding, convolutional layers, and temporal max pooling can be done in a dense fashion (retaining computational efficiency benefits), using memory proportional to what would be achieved with sparsity-exploiting code. The total memory use is now independent of the original sequence length $T$, and a diagram of the process is presented in \autoref{fig:fixed_mem_conv}. 

\label{sec:aliasing}
\textbf{Details on windowing artifacts:}
We noted that in the concatenation of different chunks in \autoref{fig:fixed_mem_conv} into one new sequence, it is technically possible for a new index to become the maximal activation due to the receptive window length $W$ crossing between two chunks that were previously not adjacent. This results in a pattern that has potentially not been seen previously, which thus creates new activation values. We have never observed this issue in practice, and so have not taken any special steps to avoid this situation (with more details in \autoref{sec:aliasing_detailed}).

This hypothetical issue could be prevented by performing the convolution and a max-pool over the chunks independently. Then, the pooled results could be concatenated and a second round of pooling performed. We have not observed any issues warranting this extra complexity and overhead.

\section{Global Channel Gating} \label{sec:global_channel_gating}

With an efficient method for handling large input sequences, we can explore a broader set of neural network architectures. In particular, we note a weakness in the original design of MalConv: the use of temporal max-pooling after a single convolutional layer results in a somewhat myopic model: learned features are purely local in nature. That is, with the existing architecture, the model output does not consider interactions between features that are far apart in time within an input sequence/file. 

To demonstrate why this is important in malware detection, consider that a common feature to learn/extract is the use of encryption libraries, which may indicate, for example, functionality common in ransomware. However, if the program does not access the file system, the use of encryption becomes less suspicious and less likely to indicate malware. In its current embodiment, it is impossible for MalConv to learn logic like this because the presence/absence of the associated information may be in disparate regions of the input, and the receptive window of the network (512 bytes) is far smaller than most inputs ($2^{21}$ bytes).

To endow our network with the ability to learn such relationships while retaining computational tractability, we develop a new attention inspired gating approach we call \textit{global channel gating} (GCG). The idea is that given a long time sequence with $C$ channels,  $X = \{\boldsymbol{x}_1, \boldsymbol{x}_2, \ldots, \boldsymbol{x}_T\}$ where $\boldsymbol{x}_{t} \in \mathbb{R}^{C}$, we want to globally suppress certain time steps based on the content of all channels. We approach this in a style similar to the gated linear unit and the additive attention \cite{DzmitryBahdana2015}, using a learned context $\boldsymbol{\bar{g}} \in \mathbb{R}^{C}$, as shown in \autoref{eq:gcg}. 

\begin{equation} \label{eq:gcg}
    GCG_{W}(\boldsymbol{x}_t,\boldsymbol{\bar{g}}) = \boldsymbol{x}_t \cdot \sigma\left(\boldsymbol{x}_t^\intercal \tanh\left(W^\intercal \boldsymbol{\bar{g}}\right)\right)
\end{equation}

The entries of the vector $\boldsymbol{x}_t \in \mathbb{R}^{C}$ at time $t$ may be suppressed by the scalar quantity on the right hand side of the GCG equation. Due to the sigmoid operation $\sigma(\cdot)$,  $\boldsymbol{x}_t$ will be scaled by a value in the range of $[0, 1]$, resulting in a context sensitive suppression of each entry in the vector. 

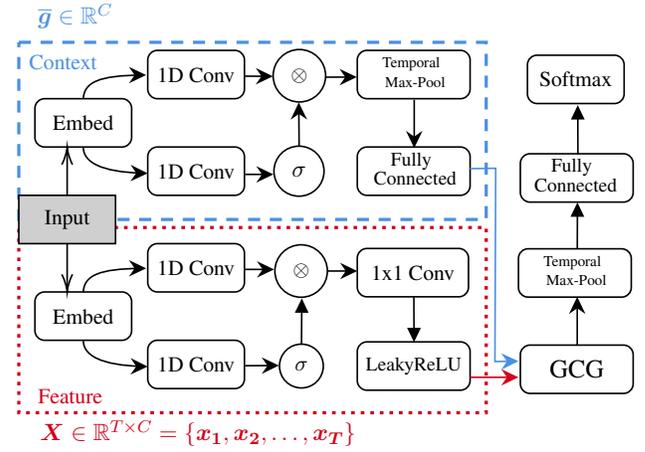
\begin{figure}[!htb]
    \centering
    \adjustbox{width=\columnwidth}{%
        \input{figs/malconv2_nocat.tex}
    }
    \caption{Our new proposed architecture with global channel gating (GCG). The \textcolor{blue}{blue} thick dashed sub-network shows the \textcolor{blue}{context extractor}, which is used to suppress information found from the \textcolor{red}{Feature extraction} sub-network (\textcolor{red}{red}, thinly dashed sub-network). 
    }
    \label{fig:malconv2}
\end{figure}

We detail a new malware classification architecture which we term \textit{MalConv with GCG} in \autoref{fig:malconv2} that leverages GCG. The top half of the network serves to learn a global context $\boldsymbol{\bar{g}}$, which is used as input to the GCG. The bottom half of the architecture shows the feature sub-network, which uses a different embedding layer to perform initial feature extraction and uses GCG to selectively suppress regions of the input, allowing for potential feature interactions over time. The inputs to GCG are a state vector from the top half context network, and a sequence over time generated from the bottom half, which has \autoref{eq:gcg} applied point-wise over time. This is followed by temporal max pooling, where we apply the fixed memory approach from \autoref{sec:fixed_memory} to make the training feasible with fixed memory costs. 

\subsection{Gating via convolution}

\begin{figure}[!h]
    \centering
    \begin{minted}[xleftmargin=0.0em,tabsize=2,breaklines,mathescape=true,fontsize=\small]{python}
def gcg(self, X, g):
  # X.shape = (B, T, C)
  B, T, C = X.size(0), X.size(1), X.size(2)
  # g.shape = (B, C)
  # create context vector $z = \tanh(W^\intercal g)$
  # self.w references a nn.Linear(C, C) layer 
  z = torch.tanh(self.w(g)) 
            
  # Size is (B, C), but we need (B, C, 1) to use as a 1d conv filter
  z = torch.unsqueeze(z, dim=2)
  # roll the batches into the channels
  x_tmp = X.view(1,B*C,-1) 
  # apply a conv with B groups; each batch gets its own context applied 
  # This computes $x_t^\intercal z$ forall $t = 1...T$
  x_tmp = F.conv1d(x_tmp, z, groups=B)
  # x_tmp has a shape of (1, B, T); re-order as (B, 1, T)
  gates = x_tmp.view(B, 1, -1)
            
  # effectively apply $x_t \cdot \sigma(x_t^\intercal \tanh(W^\intercal g))$
  return X * torch.sigmoid( gates )
    \end{minted}
    \caption{PyTorch code demonstrating how to implement global channel gating in a computationally efficient manner. The input context $g$ is projected and re-shaped, such that it can be used as the filter weights in a 1D convolution grouped by the batch size. This results in computing the dot product over time.
    }
    \label{fig:gcg_pytorch}
\end{figure}

Care must be taken to implement the GCG approach effectively. The naive strategy to implement GCG requires reshaping the input array, and either running over every time step with a for loop to extract a slice $\boldsymbol{x}_t$ and perform a dot product, or alternatively, duplicating the context $\boldsymbol{\bar{g}}$ into a larger matrix and performing a larger BLAS operation against the input $X$. The first approach suffers from excessive Python and auto-grad overhead in our testing. The latter approach is more efficient in terms of FLOPs, but still cumbersome and slow due to the duplication of $\boldsymbol{\bar{g}}$.

Instead, we exploit the nature of grouped convolutions \cite{NIPS2012_4824} to efficiently implement the GCG over time. Given a batch of $B$ time series, we reinterpret the input activation/context $\boldsymbol{\bar{g}}$ as a set of 1D convolution weights/filters in a $B \times C \times 1$ matrix, and perform a grouped convolution with $B$ groups. Thus we convolve the context with the input $X$ where the window size is 1 (considering only one time-step at a time), the $B$ different contexts become the number of output "channels", and by grouping each context is applied only to its appropriate input. The grouped convolution allows us to apply the different filters to each batch in one operation. 
We find this easiest to demonstrate with code, and present a working PyTorch implementation of GCG in \autoref{fig:gcg_pytorch}. 
With this additional insight, the GCG operation is no more expensive than a $1\times1$ convolution, allowing us to leverage it for inputs with hundreds of millions of time-steps without issue.

\section{Results} \label{sec:results}

\begin{figure}[!htb]
    \centering
    \adjustbox{width=\columnwidth}{%
        \input{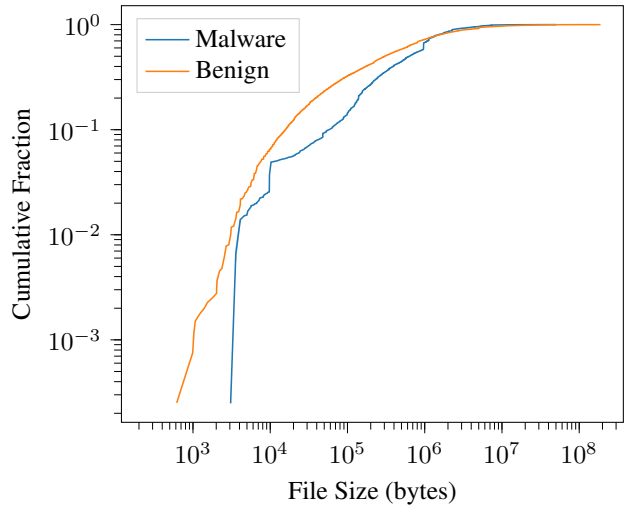}
    }
    \caption{Distribution of file lengths (x-axis, log-scale) and percentage of files of an equal or lesser size (y-axis, log-scale) for all files in the Ember2018 corpus. The largest file is 271.1 MB. }
    \label{fig:ember2018_lengths}
\end{figure}
The Ember2018 corpus \cite{Anderson2018}  has 600,000 training samples and 200,000 test samples. At $\approx$1 TB it is our primary test set due to size and difficulty. Both training and testing sets are evenly split between benign and malicious, and all testing samples were first observed after all training samples.  The predecessor 2017 corpus was explicitly noted to be "easy" due to the way it was created, and MalConv obtained an AUC of 99.8\%, close to that of a domain knowledge approach which achieved 99.9\% AUC. We prefer the 2018 corpus because it was designed to be more challenging, and MalConv obtains an accuracy of only 91\% on the newer corpus. The domain knowledge features were less impacted, dropping to only 99.6\% AUC. This better demonstrates the gap between current deep learning and domain knowledge based approaches for classifying malware. We also use the Common Crawl 
to collect  676,843 benign PDF files and VirusShare \cite{VirusShare} 158,765 malicious ones.  This gives 464 GB of data total, with 10\% used as a test set. Malicious PDF files are easier to detect than malicious executables, so the effect size of our improvements are expected to be smaller. We include this test to show that our methods still work on other types of data. 

The distribution of file lengths in bytes is shown in \autoref{fig:ember2018_lengths}, with the longest file corresponding to a time series with 271,082,368 time steps. This is $135.5\times$ longer than the original MalConv work, and thus two orders of magnitude longer than any previous time-series classification task we are aware. We were able to train Malconv with and without GCG on these data without any truncation. This removes the trivial adversarial attack of moving malicious code past the 2 MB limit.%

For all networks, we trained using the Adam optimizer \cite{Kingma2015} with the recently proposed decoupling of weight-decay \cite{Loshchilov2019}, using the recommended default parameters. A batch size of 128 was used in each experiment. All experiments were performed on a DGX-1. We note that our improved training procedure no longer requires this level of compute power, however, we do this to appropriately compare training time in our experiments with previous work. We denote MalConv trained with the original approach, truncating to the first 2MB of the input file, as ``MalConv (2MB, Orig)''. In what follows, we use "MalConv" to denote the original architecture from \autoref{fig:malconv} trained with our new fixed-memory approach specified in \autoref{sec:fixed_memory}. Finally, our new MalConv with GCG from \autoref{sec:global_channel_gating} will be the last model we train for comparison. Both MalConv and MalConv with GCG are trained to processes the entirety of the input files, up to 271 MB. We train all models for 20 epochs, using the result from the last epoch. 

For MalConv we use a filter size of 512, a stride of 512, 128 channels for each 1D Conv block, and an embedding dimension of 8. For MalConv with GCG we use a filter size of 256, a stride of 64, 256 channels for each convolution, and an embedding dimension of 8. For all models we incorporate the suggestion of \cite{Fleshman2018a} of including a special token after the EOF that maps to an embedding vector of all zeros. Details on the hyper-parameter selection, including attempts at improving the standard MalConv, can be found in \autoref{sec:param_tuning}. Below we will show the results indicating how our methods have improved MalConv, and we provide a discussion of other attempts to improve upon the MalConv approach that were unsuccessful, and how they impacted our approaches' final design in \autoref{sec:did_not_work}

\subsection{Training MalConv with fixed-memory max-pooling}

\begin{table}[!htb]
\caption{Results on training time and computational efficiency. 
} \label{tbl:low_mem_results}
\adjustbox{max width=\columnwidth}{%
\begin{tabular}{@{}lcr@{}}
\toprule
\multicolumn{1}{c}{Model} & Time Per Epoch & \multicolumn{1}{l}{GPU RAM} \\ \midrule
MalConv (2MB, Orig)       & 21 hr 29 min   & 128 GB                      \\
MalConv                   & 1 hr 10 min    & 1.1 GB                      \\
MalConv w/ GCG            & 4 hr 5 min     & 2.2 GB                      \\ \bottomrule
\end{tabular}
}
\end{table}
We first evaluate the impact of our fixed-memory approach to training over long sequences. The original MalConv required 128 GB of GPU memory, and 21.5 hours per epoch on the Ember2018 dataset. In \autoref{tbl:low_mem_results} we can see the timing information and memory use compared to our new approaches.

Our fixed-memory approach to temporal max pooling results in significant benefits, with a $116\times$ improvement in memory use and a $18.4\times$ reduction in training time. This takes MalConv training down from the order of a month to just a day. We note that the results are further improved when we consider that fixed-memory pooling is faster while processing more data, since it considers the entirety of each file. Since 14.2\% of files are greater than 2MB, we are actually processing a total $1.4\times$ more data than the original MalConv, making our speedup effectively $25.8\times$ per byte processed. Our new approach makes it possible now for anyone with a GPU and data to train MalConv.

Without these speed and memory improvements, our new MalConv with GCG architecture would not have been possible to train. Naive scaling of the results indicates we would have needed 256 GB of GPU RAM (which would have only been possible with a DGX-2), and approximately 1 month of training time.

\subsection{Improved accuracy}

\begin{table}[!htb]
\centering
\caption{Ember 2018 results on accuracy and AUC for each model. } \label{tbl:accuracy}
\adjustbox{max width=\columnwidth}{%
\begin{tabular}{lcc}
\hline
\multicolumn{1}{c}{Model} & Accuracy & AUC   \\ \hline
MalConv (2MB, Orig)       & 91.27    & 97.19 \\
MalConv                   & 91.14    & 97.29 \\
MalConv w/ GCG            & \textbf{93.29 }   & \textbf{98.04} \\ \cline{1-3}
LZJD                      & 73.43    & 84.98 \\
BWMD                      & 81.97    & 91.12 \\ \hline
\end{tabular}
}
\end{table}
In \autoref{tbl:accuracy} we show the classification performance of all three models, and two state of the art compression based methods LZJD and BWMD using 9-nearest neighbor classification. We see that training MalConv with the prior approach but on the entire sequence length has no appreciable difference in accuracy (fluctuations of 0.1 percentage points). This shows 1) that we are able to still learn effectively while processing more information, and 2) that our approach does not hinder training in any way. As noted previously, parsing all of the input file is also beneficial for thwarting the trivial attack of moving all malicious code to the end of an executable. 
We also see that our MalConv with GCG improves upon the accuracy by 2.2\% and AUC by 0.87\% of the original MalConv architecture. 

We prefer evaluation on the Ember 2018 corpus because it is both large and challenging. Our evaluations on the PDF corpus are done to show that our improvements transfer to other file types as well. On our PDF corpus we obtain an Accuracy of 99.16\% and an AUC of 99.76\%. MalConv with GCG improves this to 99.42\% and 99.80\%. Because PDF file are easier to processes, the baseline MalConv is already nearing maximal performance, so the gain is smaller --- but shows our GCG approach is still an improvement.

\subsection{Ablation testing of Avast architecture}
\begin{figure}
  \begin{center}
    \includegraphics[angle=-90,origin=c,clip=true,]{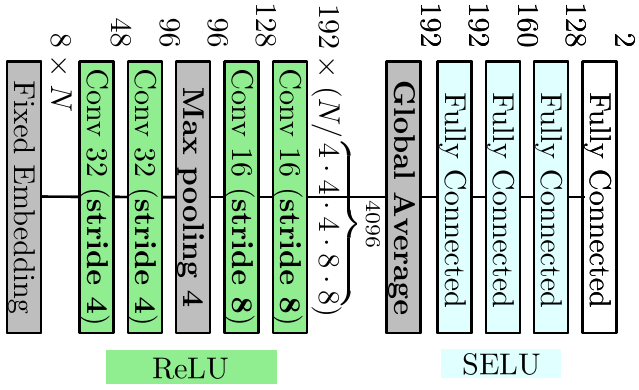}
  \end{center}
  \vspace{-37.5pt}
  \caption{
  "AvastConv" architecture from \cite{Krcal2018}.
  Their approach was originally trained on entire executables, but each executable was $\leq 512$ KB. 
  } \label{fig:avast_conv}
\end{figure}

A difficulty of research in this space is that large testing of over millions of executables can only be done in partnership with commercial anti-virus companies, who have large corpora to test against. Of the prior works we discussed in \autoref{sec:related_work}, the work by \textcite{Krcal2018} is of particular interest for two reasons, 1) they found that global max pooling produced a 20\% relative drop in performance compared to their use of global average pooling, and 2) it is the only extension to MalConv we are aware that is easy to adjust with our fixed memory max pooling form \autoref{sec:fixed_memory}. The architecture they use, which we will call \textit{AvastConv}, is given in \autoref{fig:avast_conv}. Its primary differences are the use of more layers of smaller filter widths (32 followed by 16), a hard coded embedding rather than a learned embedding, and the aforementioned use of global average pooling instead of global max pooling.

The pooling difference was the largest factor %
according to the ablation testing by \cite{Krcal2018} at 20\%. They found that fixed vs learned had no performance impact, and other differences between their and our current architectures accounted for no more than a 4\% difference. The biggest untested factor between these works is that their study was constrained to smaller executables $\leq 512 KB$ in size, where in our work we consider unbounded size with inputs over 200 MB in size.

As such, we choose to perform a small ablation test against this architecture, replacing the global average pooling with our new fixed memory temporal max pooling. Training their architecture for 20 epochs, we obtain an accuracy of 85.8\% and an AUC of 94.6\%. These results are significantly lower than MalConv and our improved version shown in \autoref{tbl:accuracy}. 

While it may be possible that global averaging would restore performance to their approach, there is not enough remaining accuracy for a 20\% relative improvement to occur. This would seem to indicate their initial results on the strength of global pooling are not as strong when factoring in larger file sizes. This is beneficial from the perspective that we can use max pooling to achieve fixed memory cost, which is not possible with average pooling. 

These results also give credence to a relatively shallower architecture with wider convolutional filters, which is maintained in our current design. This runs in contrast to normal applications of CNNs in the vision, signal, and natural language processing domains, where the community has more firmly rested on smaller filters with more layers being the canonical design approach.

\subsection{Example of interactions over time with GCG} \label{sec:interactions}

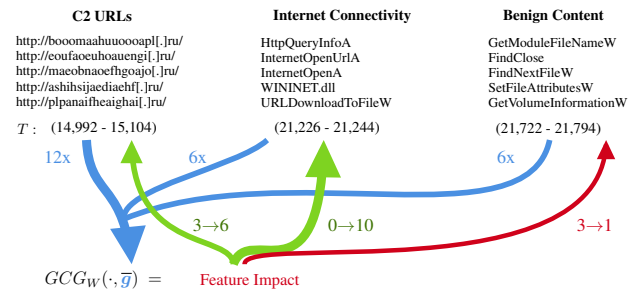
\begin{figure}[!htb]
    \centering
    \adjustbox{width=\columnwidth}{%
        \input{figs/interactionExample.tex}
    }
    \caption{The time steps $T$ that feature content is found is shown in parenthesis. 
    $\bar{g}$ \textcolor{red}{suppressed} activations from benign content, and \textcolor{blue}{increased} focus on internet functionality/  
    This shows GCG can related disconnected portions with no overlapping receptive field, learning complex interactions. 
}
    \label{fig:interaction_example}
\end{figure}

As our last result, we demonstrate an example of how our new GCG has effectively learned non-linear interactions across large time ranges. In \autoref{fig:interaction_example} we show a diagram of the relevant content from a malicious sample, and how the \textcolor{blue}{context} impacts the selected \textcolor{red}{features} once the temporal max-pooling is performed. In particular, three types of content found: Command-and-Control ("C2") URLs used for remote control of the malware, system calls used for internet connectivity, and other innocuous benign content. The \textcolor{blue}{blue} lines into the \textcolor{blue}{context} vector $\bar{g}$ denote the number of filters that had their maximum activation occur in the byte range of that content, with 12 filters selecting the C2 URLs, and 6 each for the Internet connectivity and other benign content. The values in parenthesis indicate the time-step $T$ (i.e., byte location) of the content within the file. We can see that the C2 URLs are $\geq$6,122 steps/bytes away from the rest of the content, far larger than the receptive field of the convolutions. 

After the GCG gating is applied to the activations of the \textcolor{red}{feature} sub-network, we see a large change in what is selected by the final max-pooling operation. Without GCG, none of the internet connectivity features were selected. In this case, the GCG suppresses the activations of other regions in the binary, but opens the gate (i.e., $GCG_W(x_{21,226}, \bar{g}) \approx 1.0$) for the internet content. As such, 10 of the filters now activate for this region. Similarly, the number of filters activating for the C2 URLs increases from 3 to 6. We also see that the innocuous content for working with the file system is suppressed by the gate, reducing activations from 3 down to 1. Combined, the C2 URLs and the use of APIs to connect to them over the Internet are significant indicators for confirming this file as malicious, which the network successfully performs. This is helpful information for malware analysts, and others who wish to know how the malware performs. %

This malicious examples demonstrates that our GCG mechanism successfully learns the kinds of nuanced interactions we desire over large time ranges. The use of file system and internet connectivity is intrinsically non-suspicious on their own. Correctly focusing on the right content requires observing the suspicious URLs contained withing the file. Simultaneously, this shows MalConv learning to perform sub-tasks, like determining if a URL looks suspicious or not, to make these informed contextual gating decisions.  Because this kind of analysis is expensive, we include more results from the PDF corpus in \autoref{sec:pdf_exploration}.

\section{Conclusion} \label{sec:conclusion}

Prior approaches to classifying time series of extreme length where limited by the memory required to train the models. We have developed a new approach that exploits the sparsity of temporal max pooling to make this memory cost invariant to the time series length, while simultaneously being faster to compute. Using this improvement we designed a new approach to malware detection from raw bytes using a global channel gating mechanism that gives us the capability of learning feature interactions across time, despite extreme input lengths in excess of 100 million time steps. This contribution further improves the accuracy of our malware detection model by up to 2.2\%. 
\bibliography{refs}

\clearpage

\begin{appendix}

\onecolumn

\section{What Did Not Work} \label{sec:did_not_work}

In the development of this work, we tested a number of alternative design choices. In this section we will attempt to further delve into approaches that did not work, our hypothesis as to why, and how they have influenced the final design of our approach. 

\textbf{Going Deeper:}
For both MalConv and our new GCG experiment, we also tested with up to 6 layers. In all of these experiments, using more layers never preformed better than using a single layer. This did not appear to be an issue tied to the wide filter size of our network, which we also varied from 32 to 256 (since larger ones would not consistently run without error). These results give us further pause about the nature of deep learning on extreme sequence lengths, and how to best design new architectures in this space. 

We originally desired to greatly increase the number of layers, which simply did not pan out in testing due to failure to converge or significantly poorer results. In works in the image processing domain, such as ResNet with 152 layers, we are able to create and learn higher order interactions through the depth of the network, even if two inputs within an image are not located immediately adjacent to one another. We desired to increase the depth to obtain these interaction effects. Since this was not possible, we had to devise a new method of obtaining interactions across large regions of the input, which we accomplished using our GCG approach. 

\textbf{Traditional Attention:}
Attention mechanisms have demonstrated considerable success in capturing long term dependencies among sequences \cite{Luong2015EffectiveTranslation, NIPS2017_7181, Lin2017ASS, Devlin2019BERTPO}. Unfortunately, these global attention mechanisms require a quadratic increase in memory and runtime with sequence length. We explored several applications of attention including those from document classification \cite{Yang2016HierarchicalClassification, Zhang2016Rationale-AugmentedClassification} which is the most similar NLP application to our domain. These works were limited to sequences several orders of magnitude smaller than what we require and took advantage of sub-document structure such as sentences to further reduce complexity. Current attention mechanisms remain intractable for our use case, with estimated epoch training times of at least 1 century for every method we have tried. This leads to our proposed GCG method the only viable mechanism for capturing long range dependencies in extremely long sequences with minimal additional overhead. 

\textbf{Extending PyTorch:}
As mentioned, the main idea for making MalConv faster and more efficient is to take advantage of the sparsity of the gradients caused by the Temporal Max-Pooling operation. While none of the popular deep learning frameworks take advantage of this situation natively, PyTorch does include a flexible ability to extend the framework by adding custom functions with user defined forward and backward logic. Attempting to leverage this functionality was our first approach to solving this problem.

In order to remove the unnecessary memory requirements, we attempted to merge all layers from embedding to pooling into a single custom operation. While technically feasible, re-implementing the custom logic for all of these layers while keeping only the gradients we needed became cumbersome and more complex than necessary. Furthermore, a new function would need to be created for any change in architecture. This led us to searching for the solution presented here utilizing only native PyTorch functionality, resulting in our fixed memory pooling. While retrospectively simple, developing this approach has been a roadblock to replication of the original MalConv results for several years now, as we detailed in \autoref{sec:related_work}. 

\textbf{Custom Rematerialization:}
As mentioned in \autoref{sec:related_work}, rematerialization is a common strategy to reduce memory cost. We attempted to implement custom versions of this for PyTorch that exploited the sparsity to avoid redundant computations, and to leverage the structure in global average pooling so that we could perform better ablations against the AvastConv approach. These approaches were thwarted by the increase compute of standard rematerialization. The estimated completion time for just the truncated 2MB training with rematerialization over average pooling was $\approx$5 months on our DGX1, so we abandoned the approach as beyond our computational capacity. 

\section{PDF GCG Discoveries} \label{sec:pdf_exploration}

Reverse engineering even one executable is laborious, making it costly to find examples that show the interesting things MalConv can learn. It is often the case that malware files are simple and the maliciousness obvious in a large class of files, making this a cost paid to find correct, but uninteresting, samples. They also often require more nuance technical detail that makes them undesirable to use as examples for an audience we do not expect to necessarily assembly code / reverse engineering experience.  

To help us with this situation, we have performed more work on reverse engineering the results detected by our PDF MalConv with GCG model. The results from the PDF corpus are often easier to explain, and even visual in nature, which we hope makes this more accessible to show the value of our approach.

We start by using the same approach outlined in \autoref{sec:interactions} to find regions of the input that MalConv marked as important with the GCG. 
These often occurred between the \texttt{stream} and \texttt{endstream} markers in the PDF file format, and additional processing was done to decode that data as indicated in object header for that stream. The most common encoding was \texttt{/FlateDecode}, which can easily be decoded with the \texttt{zlib} Python module. Some important images were found, which were identified by the \texttt{/DCTDecode} attribute.

\begin{figure}[H]
    \centering
    \includegraphics[scale=0.4]{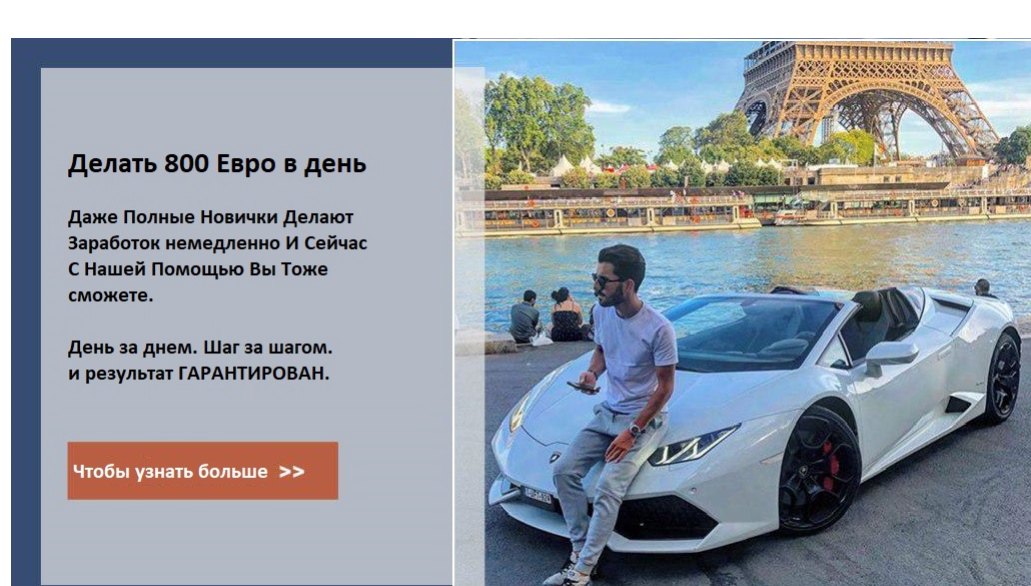}
    \caption{Make 800 Euros per day, so easy, even newbies can do it}
    \label{fig:eurosParisMan}
\end{figure}

Several images were found to be important which were obvious social engineering attempts, offering the viewer the promise of money if they would perform some activity, presumably to click a link. A lot of the images contained Russian text, possibly indicating that Russian-speaking people were the target victim group. For example, see \autoref{fig:eurosParisMan}. 

For the curios, the link in this document takes the user on a tour of abusing link redirection pages, with the series of jumps encoded as one single URL. These jump links can be abused so that when hovering the mouse over the link, the destination website might look legitimate, tricking the user into thinking the last website is also legitimate.
\begin{enumerate}
    \item www.broadcastingcable.com/common/jumplink.php?target=URL\_HERE
    \item www.creatiblogs.es/index.php?obj=front\&action=redirect\&step=1\&url=URL\_HERE
    \item c2n.me/43zBtcZ
\end{enumerate}

The final link is for a clip, screenshot, and image sharing website which is based out of Minsk, Belarus. The website and company seems legitimate, but the content there likely was not. It has since been deleted, or was removed due to expiration of content.

This is notable because it shows the MalConv with GCG mechanism identifying information across multiple file formats simultaneously, as the activation was on the Jpeg that was zip encoded within a larger PDF file. Manually accounting for this variety of format interactions is a complexity that makes feature engineering difficult, and these results show how we can help to alleviate this cost. 

\begin{figure}[H]
    \centering
    \includegraphics[scale=0.5]{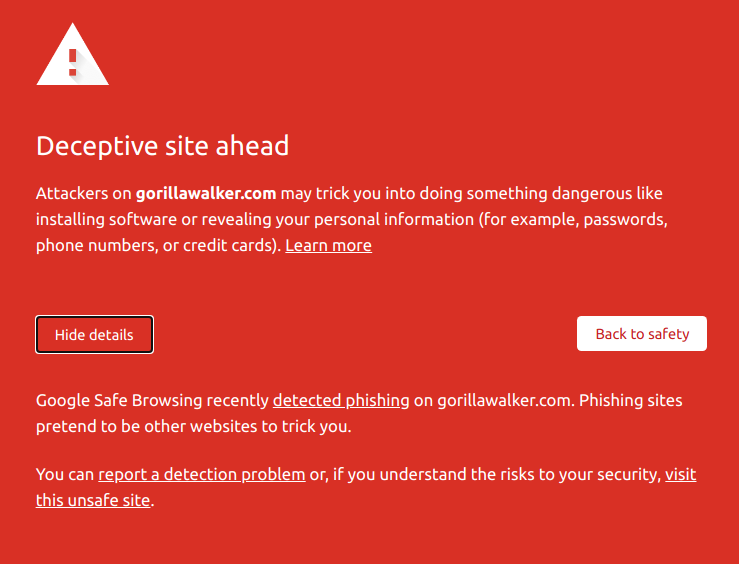}
    \caption{Visiting gorillawalker[.]com with Google Chrome 83}
    \label{fig:gorillawalkerchromewarning}
\end{figure}

Continuing on the social engineering trend, many documents had links which MalConv with GCG found to be suspicious. One of these links was GorillaWalker[.]com, a website which seems to have many PDF documents available for free download. Going to the website with Google Chrome shows a message saying that the domain is dangerous, and contains deceptive content, as shown in \autoref{fig:gorillawalkerchromewarning}. Additionally, VirusTotal can also be used to see if anti-virus or other security software have reports for malicious websites, and they certainly do for GorillaWalker. Ten out of 80 engines mark the website as Malicious, Phishing, or Spam. These many PDF document hits for this website with MalConv show that, while the URL in question is not an obvious smoking gun for malicious activity, it can lead a malware analyst to some interesting \& useful clues which might not otherwise be considered. Again, this helps show the utility and need for our GCG approach to be able to model these interactions. 

Some specific PDF malware samples which have interesting content shown by MalConv are:

\paragraph{d9823a2b316139e2b5d69d05a01ca0dd5aad644fc783a013d65a5b77e10553c4:}This sample has a TIFF image which was found by MalConv, and it turns out that TIFF image took advantage of the Acrobat TIFF renderer to achieve arbitrary code execution. The TIRR image was uploaded to Virus Total and found to be malicious.

\paragraph{000a5028b6b95abd90b7cfa9070fd8a26222c0b0ede27621cebf1ae5a20b646a:}MalConv pointed at the sequence \texttt{/MediaBox [0 0 595.280 841.890] /TrimBox [0.000 0.000 595.280 841.890]} as being malicious. It's not certain what exactly in this sequence is malice, but a quick search engine search for this string yielded search results for new malicious PDFs with the same sequence.

\paragraph{00014d95af10a46fa638a0472ef1495c28103509cd5868f8ca87ee921f95de00:} Was identified by MalConv as being malicious, and the \texttt{/OpenAction} sequence was identified as a contributing factor. This causes Adobe Acrobat to execute a block of JavaScript immediately when the document is opened, which is likely the exploit used in this malware sample.

Some PDFs had odd URLs which MalConv identified as being important for the malicious classification. Some of these URLs are:
\begin{itemize}
    \item carrisaiopdf.myhome.cx: VirusTotal marks it as benign, but shows that several malicious PDFs are hosted on the domain and several malicious PDFs reference this domain. VirusTotal also shows that the myhome.cx domain has a lot of other suspicious looking subdomains.
    \item seasasac.lflinkup.com: VirusTotal marks this domain as benign, and shows no links between this domain and malware, but VirusTotal Intelligence shows there have been a lot of people around the world inquiring about this domain since April 2019. It appears to be a repository of free eBooks.
    \item ewiiaskiopdf.linkpc.net: This sub-domain and many others have links to various spam campaigns, according to results from a quick internet search.
    \item cefasfese.4pu.com: This domain is marked as malicious on VirusTotal, with it hosting several malicious PDFs and other PDFs having links to it.
    \item cmeinasaoo.duckdns.org: This domain is also marked as malicious on VirusTotal, and VT shows that many malicious PDFs have links to this domain.
\end{itemize}

\begin{figure}[H]
    \centering
    \includegraphics[scale=0.5]{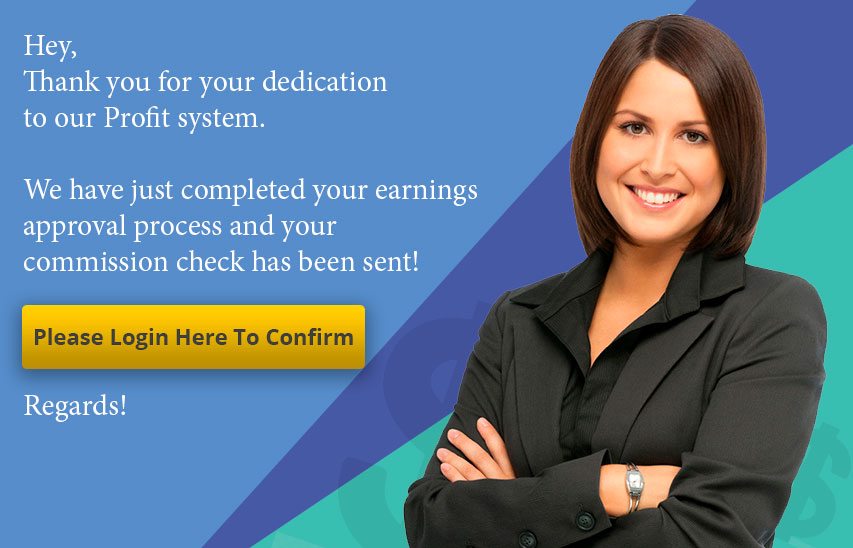}
    \caption{Just click a button to get your money}
    \label{fig:claimYourMoney}
\end{figure}

Another social engineering image is from a PDF with SHA-256 hash \texttt{0a5eeae96d4e017f8d53b1105dbe824c146e1ffe102a031ab0740e0c201e9cb6}. The image, shown in \autoref{fig:claimYourMoney}, indicates that you have a lot of money waiting for you, it's just a click away. That link would take the viewer to hxxps://apptesler2[.]com[.]cutestat[.]com, which is an expired domain, at least at the time of this writing.

\section{More Explicit Details on Windowing Artifacts} \label{sec:aliasing_detailed}

We take a moment to provide more explicit details about the issue of avoiding potential windowing artifacts discussed in \autoref{sec:aliasing}. This further exposition is done because not implementing the proposed fix actually provides certain computational benefits, which are easier to explain with some further elaboration. %

In part, we keep the current approach because if two channels, $c_1$ and $c_2$ have maximal activations that are located at time steps $t_{c_1}$ and $t_{c_2}$. If $c_1$ and $c_2$ have their maximal activation close enough within the same receptive window ($|t_{c_1}-t_{c_2}| < W/2$), we can create a smaller sub-sequence by selecting the region $|R'| \leq 2 \cdot W $ to perform the pooling over, and thus obtain faster training by having to convolve over less data. This region $R'$ would start at $\min(t_{c_1}, t_{c_2})-W/2$ and end at $\max(t_{c_1}, t_{c_2})+W/2$. 

We see this situation occur frequently due to some filters learning to look for information found within the header section of an executable. This section is usually continuous (not but required to be), and less than 400 bytes in length. As such, combining the regions results in a shorter number of bytes that need to be convolved over, while simultaneously providing greater FLOPs by performing all operations in one call to the GPU, rather then several.

As mentioned before, it is possible to completely circumvent this hypothetical issue to guarantee that the exact same activation values will always be identical. This is done by instead keeping the $C$ different chunks separate, rather than concatenating them. Perform the convolution on each chunk independently only for the valid region (i.e., do not pad the chunks) and perform max pooling on each independent chunk, giving $C$ different tensors shape $(C,1)$. Then concatenate these max-pooled results, and perform another round of max-pooling. This will guarantee that the exact same values are selected as the result of our fixed memory approach under all circumstances. This would result in additional overhead due to less computational efficiency and creating a slightly larger chain of operations to back-propagate through. 

Because we do not observe any change in the maximal channel activation using the naive approach that we have detailed, we do not find it necessary to use the more complex approach we have just described to guarantee that results are consistent. Though we present it here for completeness of the approach, and as a method to be employed should a situation or dataset occur where the naive approach is not sufficient. 

\section{Notes on Parameter Tuning Results and Environment} \label{sec:param_tuning}

We note that in attempting to obtain these results, we used the Optuna library
to perform optimization over the number of layers of convolution, filter size, stride, and learning rate for both MalConv and MalConv with GCG, and used 20\% of the training data as a held out validation set for tuning. Optuna was run with 100 trials for each model, with up to 50 epochs for training each trial. This occupied all of our GPU compute resources for several months.  
We make note of several patterns and issues we identified in performing this parameter search. 

For standard MalConv, we tested:
\begin{itemize}
\item channels $\in [32, 1024]$, exponentially distributed
\item stride $\in [4, 512]$, restricted to powers of two
\item layers $\in [1, 6]$
\item embedding dimension size $\in [4, 64]$
\end{itemize}

For MalConv with GCG, we tested:
\begin{itemize}
\item channels $\in [32, 1024]$, exponentially distributed
\item stride $\in [4, 512]$, restricted to powers of two
\item layers $\in [1, 6]$
\item embedding dimension size $\in [4, 16]$
\end{itemize}

First, we note that there was computational difficulty in performing the parameter search due to unexpected failures of CUDNN. Experiments with $\geq 512$ filters per convolution layer would intermittently result in program crashes. This issue appears to stem from large filter sizes, with large strides, over long inputs, being an unusual and rarely tested condition within most CNN implementations\footnote{We do note that a bug report had already been filed covering the issues we re-discovered. }. 
In the experiments we tested with more filters for MalConv, we found that they had no impact on the accuracy or AUC of the model. This appears to be because MalConv over-fits with $\geq 99.9\%$ training accuracy on the Ember2018 dataset, resulting in no gradient from which it can learn. As such, for MalConv our parameter tuning determined that the original parameters were just as good as any other paramter setting tested. 

For MalConv with GCG, the hyper parameter optimization settled on similar settings as MalConv, 1 layer of convolutions, with 252 channels and an embedding size of 6, and a stride of 32. For our final run over all the training data, we rounded these other parameter values to convenient sizes (252 channels to 256, embedding dimension of 6 to 8). 

In all experiments, our code used software versions of CUDA  10.1, CUDNN 7.5.1 and PyTorch 1.4.0 .

\end{appendix}

\end{document}

%% file: figs/malconv.tex
\begin{tikzpicture}










      \node (byte) [rectangle, draw=black] {Raw Byte};

      \node (embd) [rectangle, rounded corners, draw=black, right of=byte, xshift=+1.4cm] {Embedding};

      \node (conv_gate) [rectangle, rounded corners, draw=black, below right of=embd, xshift=+0.2cm, yshift=-0.2cm] {1D Conv};

      \node (conv_base) [rectangle, rounded corners, draw=black, above right of=embd, xshift=+0.2cm, yshift=+0.2cm] {1D Conv};

      \node (sigma) [circle, rounded corners, draw=black, right of=conv_gate, minimum height=0.8cm, xshift=+0.6cm] {$\sigma$};

      \node (mult) [circle, rounded corners, draw=black, right of=conv_base, minimum height=0.8cm, xshift=+0.6cm] {$\otimes$};

      \node (max_pool) [rectangle, rounded corners, draw=black, right of=mult, xshift=+1.6cm] {Temporal Max-Pooling};

      \node (flc) [rectangle, rounded corners, draw=black, below of=max_pool, yshift=+0.1cm] {Fully Connected};

      \node (out) [rectangle, rounded corners, draw=black, below of=flc, yshift=+0.1cm] {Softmax};


      \draw [thick,->,>=stealth] (byte) --  node[anchor=east] {} (embd);

      \draw [thick,->,>=stealth] (embd) --  (conv_gate);
      \draw [thick,->,>=stealth] (embd) --  (conv_base);

      \draw [thick,->,>=stealth] (conv_gate) --  (sigma);
      \draw [thick,->,>=stealth] (conv_base) --  (mult);

      \draw [thick,->,>=stealth] (sigma) --  (mult);

      \draw [thick,->,>=stealth] (mult) --  (max_pool);

      \draw [thick,->,>=stealth] (max_pool) --  (flc);

      \draw [thick,->,>=stealth] (flc) --  (out);

    \end{tikzpicture}

%% file: figs/lowMemPool.tex
\tikzset{every picture/.style={line width=0.75pt}} 

\begin{tikzpicture}[x=0.75pt,y=0.75pt,yscale=-1,xscale=1]

\draw  [color={rgb, 255:red, 80; green, 227; blue, 194 }  ,draw opacity=1 ][fill={rgb, 255:red, 80; green, 227; blue, 194 }  ,fill opacity=0.35 ] (100,40) -- (110,40) -- (110,50) -- (100,50) -- cycle ;
\draw  [color={rgb, 255:red, 189; green, 16; blue, 224 }  ,draw opacity=1 ][fill={rgb, 255:red, 189; green, 16; blue, 224 }  ,fill opacity=0.35 ] (100,50) -- (110,50) -- (110,60) -- (100,60) -- cycle ;
\draw  [color={rgb, 255:red, 80; green, 227; blue, 194 }  ,draw opacity=1 ][fill={rgb, 255:red, 80; green, 227; blue, 194 }  ,fill opacity=0.35 ] (110,40) -- (120,40) -- (120,50) -- (110,50) -- cycle ;
\draw  [color={rgb, 255:red, 80; green, 227; blue, 194 }  ,draw opacity=1 ][fill={rgb, 255:red, 80; green, 227; blue, 194 }  ,fill opacity=0.35 ] (0,40) -- (10,40) -- (10,50) -- (0,50) -- cycle ;
\draw  [color={rgb, 255:red, 80; green, 227; blue, 194 }  ,draw opacity=1 ][fill={rgb, 255:red, 80; green, 227; blue, 194 }  ,fill opacity=0.35 ] (10,40) -- (20,40) -- (20,50) -- (10,50) -- cycle ;
\draw  [color={rgb, 255:red, 80; green, 227; blue, 194 }  ,draw opacity=1 ][fill={rgb, 255:red, 80; green, 227; blue, 194 }  ,fill opacity=0.35 ] (20,40) -- (30,40) -- (30,50) -- (20,50) -- cycle ;
\draw  [color={rgb, 255:red, 80; green, 227; blue, 194 }  ,draw opacity=1 ][fill={rgb, 255:red, 80; green, 227; blue, 194 }  ,fill opacity=1 ][line width=0.75]  (30,40) -- (40,40) -- (40,50) -- (30,50) -- cycle ;
\draw  [color={rgb, 255:red, 80; green, 227; blue, 194 }  ,draw opacity=1 ][fill={rgb, 255:red, 80; green, 227; blue, 194 }  ,fill opacity=0.35 ] (40,40) -- (50,40) -- (50,50) -- (40,50) -- cycle ;
\draw  [color={rgb, 255:red, 80; green, 227; blue, 194 }  ,draw opacity=1 ][fill={rgb, 255:red, 80; green, 227; blue, 194 }  ,fill opacity=0.35 ] (50,40) -- (60,40) -- (60,50) -- (50,50) -- cycle ;
\draw  [color={rgb, 255:red, 80; green, 227; blue, 194 }  ,draw opacity=1 ][fill={rgb, 255:red, 80; green, 227; blue, 194 }  ,fill opacity=0.35 ] (60,40) -- (70,40) -- (70,50) -- (60,50) -- cycle ;
\draw  [color={rgb, 255:red, 80; green, 227; blue, 194 }  ,draw opacity=1 ][fill={rgb, 255:red, 80; green, 227; blue, 194 }  ,fill opacity=0.35 ] (70,40) -- (80,40) -- (80,50) -- (70,50) -- cycle ;
\draw  [color={rgb, 255:red, 80; green, 227; blue, 194 }  ,draw opacity=1 ][fill={rgb, 255:red, 80; green, 227; blue, 194 }  ,fill opacity=0.35 ] (80,40) -- (90,40) -- (90,50) -- (80,50) -- cycle ;
\draw  [color={rgb, 255:red, 80; green, 227; blue, 194 }  ,draw opacity=1 ][fill={rgb, 255:red, 80; green, 227; blue, 194 }  ,fill opacity=0.35 ] (90,40) -- (100,40) -- (100,50) -- (90,50) -- cycle ;
\draw  [color={rgb, 255:red, 80; green, 227; blue, 194 }  ,draw opacity=1 ][fill={rgb, 255:red, 208; green, 2; blue, 27 }  ,fill opacity=0.5 ] (120,40) -- (130,40) -- (130,50) -- (120,50) -- cycle ;
\draw  [color={rgb, 255:red, 80; green, 227; blue, 194 }  ,draw opacity=1 ][fill={rgb, 255:red, 80; green, 227; blue, 194 }  ,fill opacity=0.35 ] (130,40) -- (140,40) -- (140,50) -- (130,50) -- cycle ;
\draw  [color={rgb, 255:red, 80; green, 227; blue, 194 }  ,draw opacity=1 ][fill={rgb, 255:red, 80; green, 227; blue, 194 }  ,fill opacity=0.35 ] (140,40) -- (150,40) -- (150,50) -- (140,50) -- cycle ;
\draw  [color={rgb, 255:red, 80; green, 227; blue, 194 }  ,draw opacity=1 ][fill={rgb, 255:red, 80; green, 227; blue, 194 }  ,fill opacity=0.35 ] (150,40) -- (160,40) -- (160,50) -- (150,50) -- cycle ;
\draw  [color={rgb, 255:red, 80; green, 227; blue, 194 }  ,draw opacity=1 ][fill={rgb, 255:red, 80; green, 227; blue, 194 }  ,fill opacity=0.35 ] (160,40) -- (170,40) -- (170,50) -- (160,50) -- cycle ;
\draw  [color={rgb, 255:red, 80; green, 227; blue, 194 }  ,draw opacity=1 ][fill={rgb, 255:red, 80; green, 227; blue, 194 }  ,fill opacity=0.35 ] (170,40) -- (180,40) -- (180,50) -- (170,50) -- cycle ;
\draw  [color={rgb, 255:red, 80; green, 227; blue, 194 }  ,draw opacity=1 ][fill={rgb, 255:red, 80; green, 227; blue, 194 }  ,fill opacity=0.35 ] (180,40) -- (190,40) -- (190,50) -- (180,50) -- cycle ;
\draw  [color={rgb, 255:red, 80; green, 227; blue, 194 }  ,draw opacity=1 ][fill={rgb, 255:red, 80; green, 227; blue, 194 }  ,fill opacity=0.35 ] (190,40) -- (200,40) -- (200,50) -- (190,50) -- cycle ;
\draw  [color={rgb, 255:red, 80; green, 227; blue, 194 }  ,draw opacity=1 ][fill={rgb, 255:red, 208; green, 2; blue, 27 }  ,fill opacity=0.5 ] (200,40) -- (210,40) -- (210,50) -- (200,50) -- cycle ;
\draw  [color={rgb, 255:red, 80; green, 227; blue, 194 }  ,draw opacity=1 ][fill={rgb, 255:red, 80; green, 227; blue, 194 }  ,fill opacity=0.35 ] (210,40) -- (220,40) -- (220,50) -- (210,50) -- cycle ;
\draw  [color={rgb, 255:red, 80; green, 227; blue, 194 }  ,draw opacity=1 ][fill={rgb, 255:red, 80; green, 227; blue, 194 }  ,fill opacity=0.35 ] (220,40) -- (230,40) -- (230,50) -- (220,50) -- cycle ;
\draw  [color={rgb, 255:red, 80; green, 227; blue, 194 }  ,draw opacity=1 ][fill={rgb, 255:red, 80; green, 227; blue, 194 }  ,fill opacity=0.35 ] (230,40) -- (240,40) -- (240,50) -- (230,50) -- cycle ;
\draw  [color={rgb, 255:red, 189; green, 16; blue, 224 }  ,draw opacity=1 ][fill={rgb, 255:red, 189; green, 16; blue, 224 }  ,fill opacity=0.35 ] (0,50) -- (10,50) -- (10,60) -- (0,60) -- cycle ;
\draw  [color={rgb, 255:red, 189; green, 16; blue, 224 }  ,draw opacity=1 ][fill={rgb, 255:red, 189; green, 16; blue, 224 }  ,fill opacity=0.35 ] (10,50) -- (20,50) -- (20,60) -- (10,60) -- cycle ;
\draw  [color={rgb, 255:red, 189; green, 16; blue, 224 }  ,draw opacity=1 ][fill={rgb, 255:red, 208; green, 2; blue, 27 }  ,fill opacity=0.5 ] (20,50) -- (30,50) -- (30,60) -- (20,60) -- cycle ;
\draw  [color={rgb, 255:red, 189; green, 16; blue, 224 }  ,draw opacity=1 ][fill={rgb, 255:red, 189; green, 16; blue, 224 }  ,fill opacity=0.35 ] (30,50) -- (40,50) -- (40,60) -- (30,60) -- cycle ;
\draw  [color={rgb, 255:red, 189; green, 16; blue, 224 }  ,draw opacity=1 ][fill={rgb, 255:red, 189; green, 16; blue, 224 }  ,fill opacity=0.35 ] (40,50) -- (50,50) -- (50,60) -- (40,60) -- cycle ;
\draw  [color={rgb, 255:red, 189; green, 16; blue, 224 }  ,draw opacity=1 ][fill={rgb, 255:red, 189; green, 16; blue, 224 }  ,fill opacity=0.35 ] (50,50) -- (60,50) -- (60,60) -- (50,60) -- cycle ;
\draw  [color={rgb, 255:red, 189; green, 16; blue, 224 }  ,draw opacity=1 ][fill={rgb, 255:red, 189; green, 16; blue, 224 }  ,fill opacity=0.35 ] (60,50) -- (70,50) -- (70,60) -- (60,60) -- cycle ;
\draw  [color={rgb, 255:red, 189; green, 16; blue, 224 }  ,draw opacity=1 ][fill={rgb, 255:red, 189; green, 16; blue, 224 }  ,fill opacity=0.35 ] (70,50) -- (80,50) -- (80,60) -- (70,60) -- cycle ;
\draw  [color={rgb, 255:red, 189; green, 16; blue, 224 }  ,draw opacity=1 ][fill={rgb, 255:red, 189; green, 16; blue, 224 }  ,fill opacity=0.35 ] (80,50) -- (90,50) -- (90,60) -- (80,60) -- cycle ;
\draw  [color={rgb, 255:red, 189; green, 16; blue, 224 }  ,draw opacity=1 ][fill={rgb, 255:red, 189; green, 16; blue, 224 }  ,fill opacity=0.35 ] (90,50) -- (100,50) -- (100,60) -- (90,60) -- cycle ;
\draw  [color={rgb, 255:red, 189; green, 16; blue, 224 }  ,draw opacity=1 ][fill={rgb, 255:red, 189; green, 16; blue, 224 }  ,fill opacity=0.35 ] (110,50) -- (120,50) -- (120,60) -- (110,60) -- cycle ;
\draw  [color={rgb, 255:red, 189; green, 16; blue, 224 }  ,draw opacity=1 ][fill={rgb, 255:red, 189; green, 16; blue, 224 }  ,fill opacity=0.35 ] (120,50) -- (130,50) -- (130,60) -- (120,60) -- cycle ;
\draw  [color={rgb, 255:red, 189; green, 16; blue, 224 }  ,draw opacity=1 ][fill={rgb, 255:red, 189; green, 16; blue, 224 }  ,fill opacity=1 ] (130,50) -- (140,50) -- (140,60) -- (130,60) -- cycle ;
\draw  [color={rgb, 255:red, 189; green, 16; blue, 224 }  ,draw opacity=1 ][fill={rgb, 255:red, 189; green, 16; blue, 224 }  ,fill opacity=0.35 ] (140,50) -- (150,50) -- (150,60) -- (140,60) -- cycle ;
\draw  [color={rgb, 255:red, 189; green, 16; blue, 224 }  ,draw opacity=1 ][fill={rgb, 255:red, 189; green, 16; blue, 224 }  ,fill opacity=0.35 ] (150,50) -- (160,50) -- (160,60) -- (150,60) -- cycle ;
\draw  [color={rgb, 255:red, 189; green, 16; blue, 224 }  ,draw opacity=1 ][fill={rgb, 255:red, 189; green, 16; blue, 224 }  ,fill opacity=0.35 ] (160,50) -- (170,50) -- (170,60) -- (160,60) -- cycle ;
\draw  [color={rgb, 255:red, 189; green, 16; blue, 224 }  ,draw opacity=1 ][fill={rgb, 255:red, 189; green, 16; blue, 224 }  ,fill opacity=0.35 ] (170,50) -- (180,50) -- (180,60) -- (170,60) -- cycle ;
\draw  [color={rgb, 255:red, 189; green, 16; blue, 224 }  ,draw opacity=1 ][fill={rgb, 255:red, 189; green, 16; blue, 224 }  ,fill opacity=0.35 ] (180,50) -- (190,50) -- (190,60) -- (180,60) -- cycle ;
\draw  [color={rgb, 255:red, 189; green, 16; blue, 224 }  ,draw opacity=1 ][fill={rgb, 255:red, 208; green, 2; blue, 27 }  ,fill opacity=0.5 ] (190,50) -- (200,50) -- (200,60) -- (190,60) -- cycle ;
\draw  [color={rgb, 255:red, 189; green, 16; blue, 224 }  ,draw opacity=1 ][fill={rgb, 255:red, 189; green, 16; blue, 224 }  ,fill opacity=0.35 ] (200,50) -- (210,50) -- (210,60) -- (200,60) -- cycle ;
\draw  [color={rgb, 255:red, 189; green, 16; blue, 224 }  ,draw opacity=1 ][fill={rgb, 255:red, 189; green, 16; blue, 224 }  ,fill opacity=0.35 ] (210,50) -- (220,50) -- (220,60) -- (210,60) -- cycle ;
\draw  [color={rgb, 255:red, 189; green, 16; blue, 224 }  ,draw opacity=1 ][fill={rgb, 255:red, 189; green, 16; blue, 224 }  ,fill opacity=0.35 ] (220,50) -- (230,50) -- (230,60) -- (220,60) -- cycle ;
\draw  [color={rgb, 255:red, 189; green, 16; blue, 224 }  ,draw opacity=1 ][fill={rgb, 255:red, 189; green, 16; blue, 224 }  ,fill opacity=0.35 ] (230,50) -- (240,50) -- (240,60) -- (230,60) -- cycle ;
\draw  [color={rgb, 255:red, 245; green, 166; blue, 35 }  ,draw opacity=1 ][fill={rgb, 255:red, 245; green, 166; blue, 35 }  ,fill opacity=0.35 ] (0,60) -- (10,60) -- (10,70) -- (0,70) -- cycle ;
\draw  [color={rgb, 255:red, 245; green, 166; blue, 35 }  ,draw opacity=1 ][fill={rgb, 255:red, 245; green, 166; blue, 35 }  ,fill opacity=0.35 ] (10,60) -- (20,60) -- (20,70) -- (10,70) -- cycle ;
\draw  [color={rgb, 255:red, 245; green, 166; blue, 35 }  ,draw opacity=1 ][fill={rgb, 255:red, 245; green, 166; blue, 35 }  ,fill opacity=0.35 ] (20,60) -- (30,60) -- (30,70) -- (20,70) -- cycle ;
\draw  [color={rgb, 255:red, 245; green, 166; blue, 35 }  ,draw opacity=1 ][fill={rgb, 255:red, 208; green, 2; blue, 27 }  ,fill opacity=0.5 ] (30,60) -- (40,60) -- (40,70) -- (30,70) -- cycle ;
\draw  [color={rgb, 255:red, 245; green, 166; blue, 35 }  ,draw opacity=1 ][fill={rgb, 255:red, 245; green, 166; blue, 35 }  ,fill opacity=0.35 ] (40,60) -- (50,60) -- (50,70) -- (40,70) -- cycle ;
\draw  [color={rgb, 255:red, 245; green, 166; blue, 35 }  ,draw opacity=1 ][fill={rgb, 255:red, 245; green, 166; blue, 35 }  ,fill opacity=0.35 ] (50,60) -- (60,60) -- (60,70) -- (50,70) -- cycle ;
\draw  [color={rgb, 255:red, 245; green, 166; blue, 35 }  ,draw opacity=1 ][fill={rgb, 255:red, 245; green, 166; blue, 35 }  ,fill opacity=0.35 ] (60,60) -- (70,60) -- (70,70) -- (60,70) -- cycle ;
\draw  [color={rgb, 255:red, 245; green, 166; blue, 35 }  ,draw opacity=1 ][fill={rgb, 255:red, 245; green, 166; blue, 35 }  ,fill opacity=0.35 ] (70,60) -- (80,60) -- (80,70) -- (70,70) -- cycle ;
\draw  [color={rgb, 255:red, 245; green, 166; blue, 35 }  ,draw opacity=1 ][fill={rgb, 255:red, 208; green, 2; blue, 27 }  ,fill opacity=0.5 ] (90,60) -- (100,60) -- (100,70) -- (90,70) -- cycle ;
\draw  [color={rgb, 255:red, 245; green, 166; blue, 35 }  ,draw opacity=1 ][fill={rgb, 255:red, 245; green, 166; blue, 35 }  ,fill opacity=0.35 ] (80,60) -- (90,60) -- (90,70) -- (80,70) -- cycle ;
\draw  [color={rgb, 255:red, 245; green, 166; blue, 35 }  ,draw opacity=1 ][fill={rgb, 255:red, 245; green, 166; blue, 35 }  ,fill opacity=0.35 ] (100,60) -- (110,60) -- (110,70) -- (100,70) -- cycle ;
\draw  [color={rgb, 255:red, 245; green, 166; blue, 35 }  ,draw opacity=1 ][fill={rgb, 255:red, 245; green, 166; blue, 35 }  ,fill opacity=0.35 ] (110,60) -- (120,60) -- (120,70) -- (110,70) -- cycle ;
\draw  [color={rgb, 255:red, 245; green, 166; blue, 35 }  ,draw opacity=1 ][fill={rgb, 255:red, 245; green, 166; blue, 35 }  ,fill opacity=0.35 ] (120,60) -- (130,60) -- (130,70) -- (120,70) -- cycle ;
\draw  [color={rgb, 255:red, 245; green, 166; blue, 35 }  ,draw opacity=1 ][fill={rgb, 255:red, 245; green, 166; blue, 35 }  ,fill opacity=0.35 ] (130,60) -- (140,60) -- (140,70) -- (130,70) -- cycle ;
\draw  [color={rgb, 255:red, 245; green, 166; blue, 35 }  ,draw opacity=1 ][fill={rgb, 255:red, 245; green, 166; blue, 35 }  ,fill opacity=0.35 ] (140,60) -- (150,60) -- (150,70) -- (140,70) -- cycle ;
\draw  [color={rgb, 255:red, 245; green, 166; blue, 35 }  ,draw opacity=1 ][fill={rgb, 255:red, 245; green, 166; blue, 35 }  ,fill opacity=0.35 ] (150,60) -- (160,60) -- (160,70) -- (150,70) -- cycle ;
\draw  [color={rgb, 255:red, 245; green, 166; blue, 35 }  ,draw opacity=1 ][fill={rgb, 255:red, 245; green, 166; blue, 35 }  ,fill opacity=0.35 ] (170,60) -- (180,60) -- (180,70) -- (170,70) -- cycle ;
\draw  [color={rgb, 255:red, 245; green, 166; blue, 35 }  ,draw opacity=1 ][fill={rgb, 255:red, 245; green, 166; blue, 35 }  ,fill opacity=0.35 ] (180,60) -- (190,60) -- (190,70) -- (180,70) -- cycle ;
\draw  [color={rgb, 255:red, 245; green, 166; blue, 35 }  ,draw opacity=1 ][fill={rgb, 255:red, 245; green, 166; blue, 35 }  ,fill opacity=0.35 ] (190,60) -- (200,60) -- (200,70) -- (190,70) -- cycle ;
\draw  [color={rgb, 255:red, 245; green, 166; blue, 35 }  ,draw opacity=1 ][fill={rgb, 255:red, 245; green, 166; blue, 35 }  ,fill opacity=0.35 ] (200,60) -- (210,60) -- (210,70) -- (200,70) -- cycle ;
\draw  [color={rgb, 255:red, 245; green, 166; blue, 35 }  ,draw opacity=1 ][fill={rgb, 255:red, 245; green, 166; blue, 35 }  ,fill opacity=0.35 ] (210,60) -- (220,60) -- (220,70) -- (210,70) -- cycle ;
\draw  [color={rgb, 255:red, 245; green, 166; blue, 35 }  ,draw opacity=1 ][fill={rgb, 255:red, 245; green, 166; blue, 35 }  ,fill opacity=1 ] (220,60) -- (230,60) -- (230,70) -- (220,70) -- cycle ;
\draw  [color={rgb, 255:red, 245; green, 166; blue, 35 }  ,draw opacity=1 ][fill={rgb, 255:red, 245; green, 166; blue, 35 }  ,fill opacity=0.35 ] (230,60) -- (240,60) -- (240,70) -- (230,70) -- cycle ;
\draw    (90,30) -- (90,80) ;
\draw    (180,30) -- (180,80) ;
\draw   (89.8,27.6) .. controls (89.79,22.93) and (87.45,20.61) .. (82.78,20.62) -- (51.41,20.69) .. controls (44.74,20.7) and (41.41,18.38) .. (41.4,13.71) .. controls (41.41,18.38) and (38.08,20.72) .. (31.41,20.73)(34.41,20.73) -- (7.38,20.79) .. controls (2.71,20.8) and (0.39,23.13) .. (0.4,27.8) ;
\draw  [color={rgb, 255:red, 80; green, 227; blue, 194 }  ,draw opacity=1 ][fill={rgb, 255:red, 80; green, 227; blue, 194 }  ,fill opacity=0.35 ] (0,110) -- (10,110) -- (10,120) -- (0,120) -- cycle ;
\draw  [color={rgb, 255:red, 80; green, 227; blue, 194 }  ,draw opacity=1 ][fill={rgb, 255:red, 80; green, 227; blue, 194 }  ,fill opacity=1 ][line width=0.75]  (10,110) -- (20,110) -- (20,120) -- (10,120) -- cycle ;
\draw  [color={rgb, 255:red, 80; green, 227; blue, 194 }  ,draw opacity=1 ][fill={rgb, 255:red, 80; green, 227; blue, 194 }  ,fill opacity=0.35 ] (20,110) -- (30,110) -- (30,120) -- (20,120) -- cycle ;
\draw  [color={rgb, 255:red, 189; green, 16; blue, 224 }  ,draw opacity=1 ][fill={rgb, 255:red, 208; green, 2; blue, 27 }  ,fill opacity=0.5 ] (0,120) -- (10,120) -- (10,130) -- (0,130) -- cycle ;
\draw  [color={rgb, 255:red, 189; green, 16; blue, 224 }  ,draw opacity=1 ][fill={rgb, 255:red, 189; green, 16; blue, 224 }  ,fill opacity=0.35 ] (10,120) -- (20,120) -- (20,130) -- (10,130) -- cycle ;
\draw  [color={rgb, 255:red, 189; green, 16; blue, 224 }  ,draw opacity=1 ][fill={rgb, 255:red, 189; green, 16; blue, 224 }  ,fill opacity=0.35 ] (20,120) -- (30,120) -- (30,130) -- (20,130) -- cycle ;
\draw  [color={rgb, 255:red, 245; green, 166; blue, 35 }  ,draw opacity=1 ][fill={rgb, 255:red, 245; green, 166; blue, 35 }  ,fill opacity=0.35 ] (0,130) -- (10,130) -- (10,140) -- (0,140) -- cycle ;
\draw  [color={rgb, 255:red, 245; green, 166; blue, 35 }  ,draw opacity=1 ][fill={rgb, 255:red, 208; green, 2; blue, 27 }  ,fill opacity=0.5 ] (10,130) -- (20,130) -- (20,140) -- (10,140) -- cycle ;
\draw  [color={rgb, 255:red, 245; green, 166; blue, 35 }  ,draw opacity=1 ][fill={rgb, 255:red, 245; green, 166; blue, 35 }  ,fill opacity=0.35 ] (20,130) -- (30,130) -- (30,140) -- (20,140) -- cycle ;
\draw  [color={rgb, 255:red, 80; green, 227; blue, 194 }  ,draw opacity=1 ][fill={rgb, 255:red, 208; green, 2; blue, 27 }  ,fill opacity=0.5 ] (30,110) -- (40,110) -- (40,120) -- (30,120) -- cycle ;
\draw  [color={rgb, 255:red, 80; green, 227; blue, 194 }  ,draw opacity=1 ][fill={rgb, 255:red, 80; green, 227; blue, 194 }  ,fill opacity=0.35 ] (40,110) -- (50,110) -- (50,120) -- (40,120) -- cycle ;
\draw  [color={rgb, 255:red, 80; green, 227; blue, 194 }  ,draw opacity=1 ][fill={rgb, 255:red, 80; green, 227; blue, 194 }  ,fill opacity=0.35 ] (50,110) -- (60,110) -- (60,120) -- (50,120) -- cycle ;
\draw  [color={rgb, 255:red, 189; green, 16; blue, 224 }  ,draw opacity=1 ][fill={rgb, 255:red, 189; green, 16; blue, 224 }  ,fill opacity=0.35 ] (30,120) -- (40,120) -- (40,130) -- (30,130) -- cycle ;
\draw  [color={rgb, 255:red, 189; green, 16; blue, 224 }  ,draw opacity=1 ][fill={rgb, 255:red, 189; green, 16; blue, 224 }  ,fill opacity=1 ] (40,120) -- (50,120) -- (50,130) -- (40,130) -- cycle ;
\draw  [color={rgb, 255:red, 189; green, 16; blue, 224 }  ,draw opacity=1 ][fill={rgb, 255:red, 189; green, 16; blue, 224 }  ,fill opacity=0.35 ] (50,120) -- (60,120) -- (60,130) -- (50,130) -- cycle ;
\draw  [color={rgb, 255:red, 245; green, 166; blue, 35 }  ,draw opacity=1 ][fill={rgb, 255:red, 245; green, 166; blue, 35 }  ,fill opacity=0.35 ] (30,130) -- (40,130) -- (40,140) -- (30,140) -- cycle ;
\draw  [color={rgb, 255:red, 245; green, 166; blue, 35 }  ,draw opacity=1 ][fill={rgb, 255:red, 245; green, 166; blue, 35 }  ,fill opacity=0.35 ] (40,130) -- (50,130) -- (50,140) -- (40,140) -- cycle ;
\draw  [color={rgb, 255:red, 245; green, 166; blue, 35 }  ,draw opacity=1 ][fill={rgb, 255:red, 245; green, 166; blue, 35 }  ,fill opacity=0.35 ] (50,130) -- (60,130) -- (60,140) -- (50,140) -- cycle ;
\draw  [color={rgb, 255:red, 80; green, 227; blue, 194 }  ,draw opacity=1 ][fill={rgb, 255:red, 80; green, 227; blue, 194 }  ,fill opacity=0.35 ] (60,110) -- (70,110) -- (70,120) -- (60,120) -- cycle ;
\draw  [color={rgb, 255:red, 80; green, 227; blue, 194 }  ,draw opacity=1 ][fill={rgb, 255:red, 80; green, 227; blue, 194 }  ,fill opacity=0.35 ] (70,110) -- (80,110) -- (80,120) -- (70,120) -- cycle ;
\draw  [color={rgb, 255:red, 80; green, 227; blue, 194 }  ,draw opacity=1 ][fill={rgb, 255:red, 80; green, 227; blue, 194 }  ,fill opacity=0.35 ] (80,110) -- (90,110) -- (90,120) -- (80,120) -- cycle ;
\draw  [color={rgb, 255:red, 189; green, 16; blue, 224 }  ,draw opacity=1 ][fill={rgb, 255:red, 189; green, 16; blue, 224 }  ,fill opacity=0.35 ] (60,120) -- (70,120) -- (70,130) -- (60,130) -- cycle ;
\draw  [color={rgb, 255:red, 189; green, 16; blue, 224 }  ,draw opacity=1 ][fill={rgb, 255:red, 189; green, 16; blue, 224 }  ,fill opacity=0.35 ] (70,120) -- (80,120) -- (80,130) -- (70,130) -- cycle ;
\draw  [color={rgb, 255:red, 189; green, 16; blue, 224 }  ,draw opacity=1 ][fill={rgb, 255:red, 189; green, 16; blue, 224 }  ,fill opacity=0.35 ] (80,120) -- (90,120) -- (90,130) -- (80,130) -- cycle ;
\draw  [color={rgb, 255:red, 245; green, 166; blue, 35 }  ,draw opacity=1 ][fill={rgb, 255:red, 245; green, 166; blue, 35 }  ,fill opacity=0.35 ] (60,130) -- (70,130) -- (70,140) -- (60,140) -- cycle ;
\draw  [color={rgb, 255:red, 245; green, 166; blue, 35 }  ,draw opacity=1 ][fill={rgb, 255:red, 245; green, 166; blue, 35 }  ,fill opacity=1 ] (70,130) -- (80,130) -- (80,140) -- (70,140) -- cycle ;
\draw  [color={rgb, 255:red, 245; green, 166; blue, 35 }  ,draw opacity=1 ][fill={rgb, 255:red, 245; green, 166; blue, 35 }  ,fill opacity=0.35 ] (80,130) -- (90,130) -- (90,140) -- (80,140) -- cycle ;
\draw  [color={rgb, 255:red, 65; green, 173; blue, 148 }  ,draw opacity=1 ] (20,71.6) .. controls (19.97,75.72) and (22.02,77.79) .. (26.14,77.82) -- (26.14,77.82) .. controls (32.02,77.86) and (34.95,79.94) .. (34.92,84.05) .. controls (34.95,79.94) and (37.9,77.9) .. (43.78,77.93)(41.14,77.92) -- (43.78,77.93) .. controls (47.9,77.96) and (49.97,75.92) .. (50,71.8) ;
\draw  [color={rgb, 255:red, 65; green, 173; blue, 148 }  ,draw opacity=1 ] (29.8,109.99) .. controls (29.77,105.9) and (27.71,103.86) .. (23.62,103.89) -- (23.62,103.89) .. controls (17.78,103.93) and (14.85,101.91) .. (14.82,97.82) .. controls (14.85,101.91) and (11.94,103.97) .. (6.09,104.01)(8.72,103.99) -- (6.09,104.01) .. controls (2,104.04) and (-0.03,106.1) .. (0,110.19) ;
\draw [color={rgb, 255:red, 65; green, 173; blue, 148 }  ,draw opacity=1 ]   (35,83.18) .. controls (35,98.15) and (14.55,84.28) .. (14.8,98.78) ;

\draw  [color={rgb, 255:red, 101; green, 13; blue, 179 }  ,draw opacity=1 ] (120.4,71.6) .. controls (120.37,75.72) and (122.42,77.79) .. (126.54,77.82) -- (126.54,77.82) .. controls (132.42,77.86) and (135.35,79.94) .. (135.32,84.05) .. controls (135.35,79.94) and (138.3,77.9) .. (144.18,77.93)(141.54,77.92) -- (144.18,77.93) .. controls (148.3,77.96) and (150.37,75.92) .. (150.4,71.8) ;
\draw  [color={rgb, 255:red, 101; green, 13; blue, 179 }  ,draw opacity=1 ] (59.8,109.99) .. controls (59.83,105.9) and (57.8,103.84) .. (53.71,103.81) -- (53.71,103.81) .. controls (47.86,103.77) and (44.95,101.71) .. (44.98,97.62) .. controls (44.95,101.71) and (42.02,103.73) .. (36.18,103.7)(38.81,103.71) -- (36.18,103.7) .. controls (32.09,103.67) and (30.03,105.7) .. (30,109.79) ;
\draw [color={rgb, 255:red, 101; green, 13; blue, 179 }  ,draw opacity=1 ]   (135.4,83.38) .. controls (135.4,98.35) and (44.75,73.47) .. (45,98.22) ;

\draw  [color={rgb, 255:red, 180; green, 123; blue, 30 }  ,draw opacity=1 ] (210,71.2) .. controls (209.97,75.32) and (212.02,77.39) .. (216.14,77.42) -- (216.14,77.42) .. controls (222.02,77.46) and (224.95,79.54) .. (224.92,83.65) .. controls (224.95,79.54) and (227.9,77.5) .. (233.78,77.53)(231.14,77.52) -- (233.78,77.53) .. controls (237.9,77.56) and (239.97,75.52) .. (240,71.4) ;
\draw  [color={rgb, 255:red, 180; green, 123; blue, 30 }  ,draw opacity=1 ] (90,109.99) .. controls (89.97,105.9) and (87.91,103.86) .. (83.82,103.89) -- (83.82,103.89) .. controls (77.98,103.93) and (75.05,101.91) .. (75.02,97.82) .. controls (75.05,101.91) and (72.14,103.97) .. (66.29,104.01)(68.92,103.99) -- (66.29,104.01) .. controls (62.2,104.04) and (60.17,106.1) .. (60.2,110.19) ;
\draw [color={rgb, 255:red, 180; green, 123; blue, 30 }  ,draw opacity=1 ]   (225,83.38) .. controls (225,98.35) and (75,87.28) .. (75,98.78) ;

\draw   (90,117.5) -- (133.5,117.5) -- (133.5,110) -- (150,125) -- (133.5,140) -- (133.5,132.5) -- (90,132.5) -- cycle ;
\draw  [color={rgb, 255:red, 80; green, 227; blue, 194 }  ,draw opacity=1 ][fill={rgb, 255:red, 80; green, 227; blue, 194 }  ,fill opacity=1 ][line width=0.75]  (160,110) -- (170,110) -- (170,120) -- (160,120) -- cycle ;
\draw  [color={rgb, 255:red, 189; green, 16; blue, 224 }  ,draw opacity=1 ][fill={rgb, 255:red, 189; green, 16; blue, 224 }  ,fill opacity=1 ] (160,120) -- (170,120) -- (170,130) -- (160,130) -- cycle ;
\draw  [color={rgb, 255:red, 245; green, 166; blue, 35 }  ,draw opacity=1 ][fill={rgb, 255:red, 245; green, 166; blue, 35 }  ,fill opacity=1 ] (160,130) -- (170,130) -- (170,140) -- (160,140) -- cycle ;
\draw   (180,116) .. controls (180,112.69) and (182.69,110) .. (186,110) -- (234,110) .. controls (237.31,110) and (240,112.69) .. (240,116) -- (240,134) .. controls (240,137.31) and (237.31,140) .. (234,140) -- (186,140) .. controls (182.69,140) and (180,137.31) .. (180,134) -- cycle ;
\draw  [color={rgb, 255:red, 245; green, 166; blue, 35 }  ,draw opacity=1 ][fill={rgb, 255:red, 245; green, 166; blue, 35 }  ,fill opacity=0.35 ] (160,60) -- (170,60) -- (170,70) -- (160,70) -- cycle ;

\draw (42,9) node  [font=\small] [align=left] {{\footnotesize Chunk 1}};
\draw (21.13,58) node [anchor=south west] [inner sep=0.75pt]  [font=\scriptsize] [align=left] {{\fontfamily{pcr}\selectfont x}};
\draw (31.13,68) node [anchor=south west] [inner sep=0.75pt]  [font=\scriptsize] [align=left] {{\fontfamily{pcr}\selectfont x}};
\draw (91.13,68) node [anchor=south west] [inner sep=0.75pt]  [font=\scriptsize] [align=left] {{\fontfamily{pcr}\selectfont x}};
\draw (121.13,48) node [anchor=south west] [inner sep=0.75pt]  [font=\scriptsize] [align=left] {{\fontfamily{pcr}\selectfont x}};
\draw (191.13,58) node [anchor=south west] [inner sep=0.75pt]  [font=\scriptsize] [align=left] {{\fontfamily{pcr}\selectfont x}};
\draw (201.13,48) node [anchor=south west] [inner sep=0.75pt]  [font=\scriptsize] [align=left] {{\fontfamily{pcr}\selectfont x}};
\draw (1.13,128) node [anchor=south west] [inner sep=0.75pt]  [font=\scriptsize] [align=left] {{\fontfamily{pcr}\selectfont x}};
\draw (11.13,138) node [anchor=south west] [inner sep=0.75pt]  [font=\scriptsize] [align=left] {{\fontfamily{pcr}\selectfont x}};
\draw (31.13,118) node [anchor=south west] [inner sep=0.75pt]  [font=\scriptsize] [align=left] {{\fontfamily{pcr}\selectfont x}};
\draw (112.5,127) node  [font=\scriptsize] [align=left] {MaxPool};
\draw (210.5,119.5) node   [align=left] {{\tiny FC Layers }};
\draw (210.5,129.5) node   [align=left] {{\tiny \& Softmax}};

\end{tikzpicture}

%% file: figs/malconv2_nocat.tex
\tikzset{every picture/.style={line width=0.75pt}} 

\begin{tikzpicture}[x=0.75pt,y=0.75pt,yscale=-1,xscale=1]

\draw  [color={rgb, 255:red, 208; green, 2; blue, 27 }  ,draw opacity=1 ][dash pattern={on 1.69pt off 2.76pt}][line width=1.5]  (8.82,142.5) -- (298.82,142.5) -- (298.82,257.5) -- (8.82,257.5) -- cycle ;
\draw  [color={rgb, 255:red, 74; green, 144; blue, 226 }  ,draw opacity=1 ][dash pattern={on 5.63pt off 4.5pt}][line width=1.5]  (8.82,27.5) -- (298.82,27.5) -- (298.82,137.5) -- (8.82,137.5) -- cycle ;
\draw   (88.82,38.5) .. controls (88.82,35.19) and (91.51,32.5) .. (94.82,32.5) -- (142.82,32.5) .. controls (146.13,32.5) and (148.82,35.19) .. (148.82,38.5) -- (148.82,56.5) .. controls (148.82,59.81) and (146.13,62.5) .. (142.82,62.5) -- (94.82,62.5) .. controls (91.51,62.5) and (88.82,59.81) .. (88.82,56.5) -- cycle ;
\draw   (88.82,98.5) .. controls (88.82,95.19) and (91.51,92.5) .. (94.82,92.5) -- (142.82,92.5) .. controls (146.13,92.5) and (148.82,95.19) .. (148.82,98.5) -- (148.82,116.5) .. controls (148.82,119.81) and (146.13,122.5) .. (142.82,122.5) -- (94.82,122.5) .. controls (91.51,122.5) and (88.82,119.81) .. (88.82,116.5) -- cycle ;
\draw   (18.82,68.5) .. controls (18.82,65.19) and (21.51,62.5) .. (24.82,62.5) -- (72.82,62.5) .. controls (76.13,62.5) and (78.82,65.19) .. (78.82,68.5) -- (78.82,86.5) .. controls (78.82,89.81) and (76.13,92.5) .. (72.82,92.5) -- (24.82,92.5) .. controls (21.51,92.5) and (18.82,89.81) .. (18.82,86.5) -- cycle ;
\draw   (218.82,38.5) .. controls (218.82,35.19) and (221.51,32.5) .. (224.82,32.5) -- (282.82,32.5) .. controls (286.13,32.5) and (288.82,35.19) .. (288.82,38.5) -- (288.82,56.5) .. controls (288.82,59.81) and (286.13,62.5) .. (282.82,62.5) -- (224.82,62.5) .. controls (221.51,62.5) and (218.82,59.81) .. (218.82,56.5) -- cycle ;
\draw   (218.82,98.5) .. controls (218.82,95.19) and (221.51,92.5) .. (224.82,92.5) -- (282.82,92.5) .. controls (286.13,92.5) and (288.82,95.19) .. (288.82,98.5) -- (288.82,116.5) .. controls (288.82,119.81) and (286.13,122.5) .. (282.82,122.5) -- (224.82,122.5) .. controls (221.51,122.5) and (218.82,119.81) .. (218.82,116.5) -- cycle ;
\draw    (48.82,92.83) .. controls (49.14,101.84) and (59.24,107.98) .. (85.86,108.77) ;
\draw [shift={(88.82,108.83)}, rotate = 180.83] [fill={rgb, 255:red, 0; green, 0; blue, 0 }  ][line width=0.08]  [draw opacity=0] (10.72,-5.15) -- (0,0) -- (10.72,5.15) -- (7.12,0) -- cycle    ;

\draw    (48.82,62.75) .. controls (49.3,52.13) and (58.63,48.12) .. (84.9,47.85) ;
\draw [shift={(87.82,47.83)}, rotate = 180] [fill={rgb, 255:red, 0; green, 0; blue, 0 }  ][line width=0.08]  [draw opacity=0] (10.72,-5.15) -- (0,0) -- (10.72,5.15) -- (7.12,0) -- cycle    ;

\draw    (148.82,48.5) -- (163.51,48.59) ;
\draw [shift={(166.51,48.61)}, rotate = 180.36] [fill={rgb, 255:red, 0; green, 0; blue, 0 }  ][line width=0.08]  [draw opacity=0] (10.72,-5.15) -- (0,0) -- (10.72,5.15) -- (7.12,0) -- cycle    ;

\draw    (199.44,49.13) -- (215.51,49.07) ;
\draw [shift={(218.51,49.06)}, rotate = 539.79] [fill={rgb, 255:red, 0; green, 0; blue, 0 }  ][line width=0.08]  [draw opacity=0] (10.72,-5.15) -- (0,0) -- (10.72,5.15) -- (7.12,0) -- cycle    ;

\draw    (149.18,109.5) -- (163.82,109.5) ;
\draw [shift={(166.82,109.5)}, rotate = 180] [fill={rgb, 255:red, 0; green, 0; blue, 0 }  ][line width=0.08]  [draw opacity=0] (10.72,-5.15) -- (0,0) -- (10.72,5.15) -- (7.12,0) -- cycle    ;

\draw    (182.74,91.72) -- (182.54,67.83) ;
\draw [shift={(182.51,64.83)}, rotate = 449.53] [fill={rgb, 255:red, 0; green, 0; blue, 0 }  ][line width=0.08]  [draw opacity=0] (10.72,-5.15) -- (0,0) -- (10.72,5.15) -- (7.12,0) -- cycle    ;

\draw    (254.15,63.17) -- (254.15,90.17) ;
\draw [shift={(254.15,93.17)}, rotate = 270] [fill={rgb, 255:red, 0; green, 0; blue, 0 }  ][line width=0.08]  [draw opacity=0] (10.72,-5.15) -- (0,0) -- (10.72,5.15) -- (7.12,0) -- cycle    ;

\draw   (88.82,158.5) .. controls (88.82,155.19) and (91.51,152.5) .. (94.82,152.5) -- (142.82,152.5) .. controls (146.13,152.5) and (148.82,155.19) .. (148.82,158.5) -- (148.82,176.5) .. controls (148.82,179.81) and (146.13,182.5) .. (142.82,182.5) -- (94.82,182.5) .. controls (91.51,182.5) and (88.82,179.81) .. (88.82,176.5) -- cycle ;
\draw   (88.82,218.5) .. controls (88.82,215.19) and (91.51,212.5) .. (94.82,212.5) -- (142.82,212.5) .. controls (146.13,212.5) and (148.82,215.19) .. (148.82,218.5) -- (148.82,236.5) .. controls (148.82,239.81) and (146.13,242.5) .. (142.82,242.5) -- (94.82,242.5) .. controls (91.51,242.5) and (88.82,239.81) .. (88.82,236.5) -- cycle ;
\draw   (18.82,188.5) .. controls (18.82,185.19) and (21.51,182.5) .. (24.82,182.5) -- (72.82,182.5) .. controls (76.13,182.5) and (78.82,185.19) .. (78.82,188.5) -- (78.82,206.5) .. controls (78.82,209.81) and (76.13,212.5) .. (72.82,212.5) -- (24.82,212.5) .. controls (21.51,212.5) and (18.82,209.81) .. (18.82,206.5) -- cycle ;
\draw   (218.82,219.5) .. controls (218.82,216.19) and (221.51,213.5) .. (224.82,213.5) -- (282.82,213.5) .. controls (286.13,213.5) and (288.82,216.19) .. (288.82,219.5) -- (288.82,237.5) .. controls (288.82,240.81) and (286.13,243.5) .. (282.82,243.5) -- (224.82,243.5) .. controls (221.51,243.5) and (218.82,240.81) .. (218.82,237.5) -- cycle ;
\draw   (319.99,102.7) .. controls (319.99,99.39) and (322.67,96.7) .. (325.99,96.7) -- (383.99,96.7) .. controls (387.3,96.7) and (389.99,99.39) .. (389.99,102.7) -- (389.99,120.7) .. controls (389.99,124.01) and (387.3,126.7) .. (383.99,126.7) -- (325.99,126.7) .. controls (322.67,126.7) and (319.99,124.01) .. (319.99,120.7) -- cycle ;
\draw    (48.82,212.83) .. controls (49.14,221.84) and (59.24,227.98) .. (85.86,228.77) ;
\draw [shift={(88.82,228.83)}, rotate = 180.83] [fill={rgb, 255:red, 0; green, 0; blue, 0 }  ][line width=0.08]  [draw opacity=0] (10.72,-5.15) -- (0,0) -- (10.72,5.15) -- (7.12,0) -- cycle    ;

\draw    (48.82,182.75) .. controls (49.3,172.14) and (58.63,168.12) .. (84.9,167.85) ;
\draw [shift={(87.82,167.83)}, rotate = 180] [fill={rgb, 255:red, 0; green, 0; blue, 0 }  ][line width=0.08]  [draw opacity=0] (10.72,-5.15) -- (0,0) -- (10.72,5.15) -- (7.12,0) -- cycle    ;

\draw    (148.82,168.5) -- (164.22,168.5) ;
\draw [shift={(167.22,168.5)}, rotate = 180] [fill={rgb, 255:red, 0; green, 0; blue, 0 }  ][line width=0.08]  [draw opacity=0] (10.72,-5.15) -- (0,0) -- (10.72,5.15) -- (7.12,0) -- cycle    ;

\draw    (200.42,169.5) -- (215.82,169.5) ;
\draw [shift={(218.82,169.5)}, rotate = 180] [fill={rgb, 255:red, 0; green, 0; blue, 0 }  ][line width=0.08]  [draw opacity=0] (10.72,-5.15) -- (0,0) -- (10.72,5.15) -- (7.12,0) -- cycle    ;

\draw    (149.18,228.3) -- (167.22,228.2) ;
\draw [shift={(170.22,228.18)}, rotate = 539.6700000000001] [fill={rgb, 255:red, 0; green, 0; blue, 0 }  ][line width=0.08]  [draw opacity=0] (10.72,-5.15) -- (0,0) -- (10.72,5.15) -- (7.12,0) -- cycle    ;

\draw    (184.2,215.32) -- (184.4,189.5) ;
\draw [shift={(184.42,186.5)}, rotate = 450.43] [fill={rgb, 255:red, 0; green, 0; blue, 0 }  ][line width=0.08]  [draw opacity=0] (8.93,-4.29) -- (0,0) -- (8.93,4.29) -- cycle    ;

\draw  [fill={rgb, 255:red, 207; green, 207; blue, 207 }  ,fill opacity=1 ] (8.82,122.5) -- (68.82,122.5) -- (68.82,152.5) -- (8.82,152.5) -- cycle ;
\draw   (318.82,161.7) .. controls (318.82,158.39) and (321.51,155.7) .. (324.82,155.7) -- (382.82,155.7) .. controls (386.13,155.7) and (388.82,158.39) .. (388.82,161.7) -- (388.82,179.7) .. controls (388.82,183.01) and (386.13,185.7) .. (382.82,185.7) -- (324.82,185.7) .. controls (321.51,185.7) and (318.82,183.01) .. (318.82,179.7) -- cycle ;
\draw    (355.41,129.9) -- (355.32,155.63) ;

\draw [shift={(355.42,126.9)}, rotate = 90.2] [fill={rgb, 255:red, 0; green, 0; blue, 0 }  ][line width=0.08]  [draw opacity=0] (8.93,-4.29) -- (0,0) -- (8.93,4.29) -- cycle    ;
\draw   (320,221.5) .. controls (320,218.19) and (322.69,215.5) .. (326,215.5) -- (384,215.5) .. controls (387.31,215.5) and (390,218.19) .. (390,221.5) -- (390,239.5) .. controls (390,242.81) and (387.31,245.5) .. (384,245.5) -- (326,245.5) .. controls (322.69,245.5) and (320,242.81) .. (320,239.5) -- cycle ;
\draw    (355.29,215.5) -- (355.29,188.5) ;
\draw [shift={(355.29,185.5)}, rotate = 450] [fill={rgb, 255:red, 0; green, 0; blue, 0 }  ][line width=0.08]  [draw opacity=0] (10.72,-5.15) -- (0,0) -- (10.72,5.15) -- (7.12,0) -- cycle    ;

\draw    (38.82,152.5) -- (38.82,180.5) ;
\draw [shift={(38.82,182.5)}, rotate = 270] [color={rgb, 255:red, 0; green, 0; blue, 0 }  ][line width=0.75]    (10.93,-3.29) .. controls (6.95,-1.4) and (3.31,-0.3) .. (0,0) .. controls (3.31,0.3) and (6.95,1.4) .. (10.93,3.29)   ;

\draw    (38.82,122.5) -- (38.82,94.5) ;
\draw [shift={(38.82,92.5)}, rotate = 450] [color={rgb, 255:red, 0; green, 0; blue, 0 }  ][line width=0.75]    (10.93,-3.29) .. controls (6.95,-1.4) and (3.31,-0.3) .. (0,0) .. controls (3.31,0.3) and (6.95,1.4) .. (10.93,3.29)   ;

\draw [color={rgb, 255:red, 208; green, 2; blue, 27 }  ,draw opacity=1 ]   (288.82,235.5) -- (315.82,235.5) ;
\draw [shift={(318.82,235.5)}, rotate = 180] [fill={rgb, 255:red, 208; green, 2; blue, 27 }  ,fill opacity=1 ][line width=0.08]  [draw opacity=0] (10.72,-5.15) -- (0,0) -- (10.72,5.15) -- (7.12,0) -- cycle    ;

\draw   (324.39,41.5) .. controls (324.39,38.19) and (327.07,35.5) .. (330.39,35.5) -- (378.39,35.5) .. controls (381.7,35.5) and (384.39,38.19) .. (384.39,41.5) -- (384.39,59.5) .. controls (384.39,62.81) and (381.7,65.5) .. (378.39,65.5) -- (330.39,65.5) .. controls (327.07,65.5) and (324.39,62.81) .. (324.39,59.5) -- cycle ;
\draw [color={rgb, 255:red, 74; green, 144; blue, 226 }  ,draw opacity=1 ]   (288.71,105.92) -- (304.71,105.7) -- (304.22,225.7) -- (315.82,225.54) ;
\draw [shift={(318.82,225.5)}, rotate = 539.22] [fill={rgb, 255:red, 74; green, 144; blue, 226 }  ,fill opacity=1 ][line width=0.08]  [draw opacity=0] (10.72,-5.15) -- (0,0) -- (10.72,5.15) -- (7.12,0) -- cycle    ;

\draw    (355.72,96.9) -- (355.63,69.5) ;
\draw [shift={(355.62,66.5)}, rotate = 449.81] [fill={rgb, 255:red, 0; green, 0; blue, 0 }  ][line width=0.08]  [draw opacity=0] (8.93,-4.29) -- (0,0) -- (8.93,4.29) -- cycle    ;

\draw   (218.82,161.5) .. controls (218.82,158.19) and (221.51,155.5) .. (224.82,155.5) -- (281.82,155.5) .. controls (285.13,155.5) and (287.82,158.19) .. (287.82,161.5) -- (287.82,179.5) .. controls (287.82,182.81) and (285.13,185.5) .. (281.82,185.5) -- (224.82,185.5) .. controls (221.51,185.5) and (218.82,182.81) .. (218.82,179.5) -- cycle ;
\draw    (254.03,210.7) -- (253.93,185.72) ;

\draw [shift={(254.04,213.7)}, rotate = 269.77] [fill={rgb, 255:red, 0; green, 0; blue, 0 }  ][line width=0.08]  [draw opacity=0] (8.93,-4.29) -- (0,0) -- (8.93,4.29) -- cycle    ;

\draw (48.82,77.5) node   [align=left] {Embed};
\draw (118.82,47.5) node   [align=left] {1D Conv};
\draw (118.82,107.5) node   [align=left] {1D Conv};
\draw (252.32,41) node   [align=left] {{\scriptsize Temporal}};
\draw (254.82,52) node   [align=left] {{\scriptsize Max-Pool}};
\draw (252.32,101) node   [align=left] {{\footnotesize Fully}};
\draw (254.82,112) node   [align=left] {{\footnotesize Connected}};
\draw  [line width=0.75]   (182.65, 107.5) circle [x radius= 15.53, y radius= 15.53]   ;
\draw (182.65,107.5) node  [font=\large]  {$\sigma $};
\draw  [line width=0.75]   (182.9, 48.25) circle [x radius= 16.28, y radius= 16.28]   ;
\draw (182.9,48.25) node  [font=\normalsize]  {$\otimes $};
\draw (48.82,197.5) node   [align=left] {Embed};
\draw (118.82,167.5) node   [align=left] {1D Conv};
\draw (118.82,227.5) node   [align=left] {1D Conv};
\draw (351.82,105.2) node   [align=left] {{\footnotesize Fully}};
\draw (354.32,116.2) node   [align=left] {{\footnotesize Connected}};
\draw  [line width=0.75]   (183.99, 169.5) circle [x radius= 16.28, y radius= 16.28]   ;
\draw (183.99,169.5) node  [font=\normalsize]  {${\textstyle \otimes }$};
\draw  [line width=0.75]   (184.15, 228.5) circle [x radius= 13.6, y radius= 13.6]   ;
\draw (184.15,228.5) node  [font=\normalsize]  {$\sigma $};
\draw (38.82,137.5) node   [align=left] {Input};
\draw (352.32,164.2) node   [align=left] {{\scriptsize Temporal}};
\draw (354.82,175.2) node   [align=left] {{\scriptsize Max-Pool}};
\draw (355,230.5) node   [align=left] {{\Large GCG}};
\draw (354.39,50.5) node   [align=left] {Softmax};
\draw (36.32,39.5) node  [font=\normalsize,color={rgb, 255:red, 74; green, 144; blue, 226 }  ,opacity=1 ] [align=left] {Context};
\draw (40.82,247) node  [color={rgb, 255:red, 208; green, 2; blue, 27 }  ,opacity=1 ] [align=left] {Feature};
\draw (120,270.17) node  [font=\large,color={rgb, 255:red, 208; green, 2; blue, 27 }  ,opacity=1 ]  {$\boldsymbol{X} \in \mathbb{R}^{T\times C} =\{\boldsymbol{x_{1}} ,\boldsymbol{x_{2}} ,\dotsc ,\boldsymbol{x_{T}}\}$};
\draw (44,11) node  [font=\large,color={rgb, 255:red, 74; green, 144; blue, 226 }  ,opacity=1 ]  {$\boldsymbol{\overline{g}} \in \mathbb{R}^{C}$};
\draw (253.32,170.5) node   [align=left] {1x1 Conv};
\draw (253.82,228.5) node   [align=left] {{\footnotesize LeakyReLU}};

\end{tikzpicture}

%% file: figs/interactionExample.tex
\tikzset{every picture/.style={line width=0.75pt}} 

\begin{tikzpicture}[x=0.75pt,y=0.75pt,yscale=-1,xscale=1]

\draw [color={rgb, 255:red, 74; green, 144; blue, 226 }  ,draw opacity=1 ][line width=6]    (105.71,219.42) .. controls (96.57,156.53) and (69.38,169.4) .. (67,120) ;

\draw [shift={(107,230)}, rotate = 264.29] [fill={rgb, 255:red, 74; green, 144; blue, 226 }  ,fill opacity=1 ][line width=0.08]  [draw opacity=0] (33.04,-15.87) -- (0,0) -- (33.04,15.87) -- cycle    ;
\draw [color={rgb, 255:red, 74; green, 144; blue, 226 }  ,draw opacity=1 ][line width=3.75]    (100,190) .. controls (98.5,162) and (200.5,144) .. (227,120) ;

\draw [color={rgb, 255:red, 74; green, 144; blue, 226 }  ,draw opacity=1 ][line width=3.75]    (100,190) .. controls (242.5,132) and (457.5,213) .. (477,120) ;

\draw [color={rgb, 255:red, 126; green, 211; blue, 33 }  ,draw opacity=1 ][line width=3.75]    (197,230) .. controls (195.53,202.56) and (110.02,220.26) .. (107.08,125.92) ;
\draw [shift={(107,120)}, rotate = 450.28] [fill={rgb, 255:red, 126; green, 211; blue, 33 }  ,fill opacity=1 ][line width=0.08]  [draw opacity=0] (20.54,-9.87) -- (0,0) -- (20.54,9.87) -- cycle    ;

\draw [color={rgb, 255:red, 126; green, 211; blue, 33 }  ,draw opacity=1 ][line width=6]    (197,230) .. controls (211.21,191.78) and (272,266.89) .. (276.78,128.73) ;
\draw [shift={(277,120)}, rotate = 450.95] [fill={rgb, 255:red, 126; green, 211; blue, 33 }  ,fill opacity=1 ][line width=0.08]  [draw opacity=0] (33.04,-15.87) -- (0,0) -- (33.04,15.87) -- cycle    ;

\draw [color={rgb, 255:red, 208; green, 2; blue, 27 }  ,draw opacity=1 ][line width=3]    (207,230) .. controls (205.52,202.28) and (520.11,255.91) .. (526.89,124.05) ;
\draw [shift={(527,120)}, rotate = 450.21] [fill={rgb, 255:red, 208; green, 2; blue, 27 }  ,fill opacity=1 ][line width=0.08]  [draw opacity=0] (16.97,-8.15) -- (0,0) -- (16.97,8.15) -- cycle    ;

\draw (78.5,60) node  [font=\footnotesize] [align=left] {http://booomaahuuoooapl[.]ru/\\http://eoufaoeuhoauengi[.]ru/\\http://maeobnaoefhgoajo[.]ru/\\http://ashihsijaediaehf[.]ru/\\http://plpanaifheaighai[.]ru/};
\draw (82.5,10.5) node   [align=left] {\textbf{C2 URLs }};
\draw (279.5,60) node  [font=\footnotesize] [align=left] {HttpQueryInfoA\\InternetOpenUrlA\\InternetOpenA\\WININET.dll\\URLDownloadToFileW};
\draw (293.5,10.5) node   [align=left] {\textbf{Internet Connectivity}};
\draw (484.5,60) node  [font=\footnotesize] [align=left] {GetModuleFileNameW\\FindClose\\FindNextFileW\\SetFileAttributesW\\GetVolumeInformationW};
\draw (480.5,10.5) node   [align=left] {\textbf{Benign Content}};
\draw (280,109.5) node   [align=left] {(21,226 - 21,244)};
\draw (82.5,109.5) node   [align=left] {(14,992 - 15,104)};
\draw (476,110.5) node   [align=left] {(21,722 - 21,794)};
\draw (82,243.5) node  [font=\Large]  {$GCG_{W}( \cdot ,\overline{\textcolor[rgb]{0.29,0.56,0.89}{\boldsymbol{g}}}) \ =$};
\draw (41.5,135) node  [font=\Large,color={rgb, 255:red, 0; green, 0; blue, 0 }  ,opacity=1 ] [align=left] {\textcolor[rgb]{0.29,0.56,0.89}{12x}};
\draw (165.5,135) node  [font=\Large,color={rgb, 255:red, 0; green, 0; blue, 0 }  ,opacity=1 ] [align=left] {\textcolor[rgb]{0.29,0.56,0.89}{6x}};
\draw (438.5,135) node  [font=\Large,color={rgb, 255:red, 0; green, 0; blue, 0 }  ,opacity=1 ] [align=left] {\textcolor[rgb]{0.29,0.56,0.89}{6x}};
\draw (211.5,245.5) node  [font=\large] [align=left] {\textcolor[rgb]{0.82,0.01,0.11}{Feature Impact}};
\draw (177,195) node  [font=\Large] [align=left] {\textcolor[rgb]{0.25,0.46,0.02}{3$\rightarrow$6}};
\draw (301,195) node  [font=\Large] [align=left] {\textcolor[rgb]{0.25,0.46,0.02}{0$\rightarrow$10}};
\draw (517,195) node  [font=\Large] [align=left] {\textcolor[rgb]{0.82,0.01,0.11}{3$\rightarrow$1}};
\draw (14.5,110) node    {$T:$};

\end{tikzpicture}